\documentclass[runningheads]{llncs}

\usepackage{eccv}

\usepackage{eccvabbrv}

\usepackage{graphicx}
\usepackage{booktabs}

\usepackage{kotex}
\usepackage{multirow}
\usepackage{amsfonts}

\usepackage[accsupp]{axessibility}

\usepackage{hyperref}

\usepackage{orcidlink}

\begin{document}

\title{Nickel and Diming Your GAN: A Dual-Method Approach to Enhancing GAN Efficiency via Knowledge Distillation} 

\titlerunning{Nickel and Diming Your GAN}

\author{Sangyeop Yeo\orcidlink{0000-0002-5305-3443} \and
Yoojin Jang\orcidlink{0000-0001-8150-3715} \and
Jaejun Yoo\textsuperscript{*}\orcidlink{0000-0001-5252-9668}}

\authorrunning{S. Yeo et al.}

\institute{Laboratory of Advanced Imaging Technology (LAIT) \\Ulsan National Institute of Science and Technology (UNIST) \email{sangyeop377@gmail.com, \{softjin, jaejun.yoo\}@unist.ac.kr \\(*: corresponding author)}}

\maketitle

\begin{abstract}
  In this paper, we address the challenge of compressing generative adversarial networks (GANs) for deployment in resource-constrained environments by proposing two novel methods:  Distribution Matching for Efficient compression (\texttt{DiME}) and Network Interactive Compression via Knowledge Exchange and Learning (\texttt{NICKEL}). \texttt{DiME} employs foundation models as embedding kernels for efficient distribution matching, leveraging maximum mean discrepancy to facilitate effective knowledge distillation. \texttt{NICKEL} employs an interactive compression method that enhances the communication between the student generator and discriminator, achieving a balanced and stable compression process. Our comprehensive evaluation on the StyleGAN2 architecture with the FFHQ dataset shows the effectiveness of our approach, with \texttt{NICKEL \& DiME} achieving FID scores of 10.45 and 15.93 at compression rates of 95.73\% and 98.92\%, respectively. Remarkably, our methods sustain generative quality even at an extreme compression rate of 99.69\%, surpassing the previous state-of-the-art performance by a large margin. These findings not only show our methodologies' capacity to significantly lower GANs' computational demands but also pave the way for deploying high-quality GAN models in settings with limited resources. Our code is available at \href{https://lait-cvlab.github.io/Nickel_and_Diming_Your_GAN/}{Nickel \& Dime}.
\keywords{Model compression \and Generative models \and Compact models}
\end{abstract}

\section{Introduction}
\label{sec:intro}
Generative Adversarial Networks (GANs) have attracted significant popularity as one of the most promising generative models, alongside the diffusion models \cite{ho2020denoising, song2020score, dhariwal2021diffusion, rombach2022high}, in various computer vision tasks such as super-resolution \cite{park2023content, ledig2017photo, wang2018esrgan}, image editing \cite{harkonen2020ganspace, shen2020interfacegan, patashnik2021styleclip}, and image generation \cite{karras2019style, karras2020analyzing, kang2023scaling}. Particularly, thanks to their fast inference speed compared to diffusion models, GANs offer significant advantages for real-time applications\cite{rippel2017real, hu2019rtsrgan, kim2021exploiting}. However, despite their outstanding performance, the application of state-of-the-art GANs~\cite{goodfellow2014generative, karras2017progressive, karras2019style, karras2020analyzing, sauer2022stylegan, sauer2023stylegan, kang2023scaling} on edge devices is constrained by their huge resource consumption.

\begin{figure}[tb!]
  \centering
  \begin{subfigure}{0.32\linewidth}
    \includegraphics[height=3.3cm]{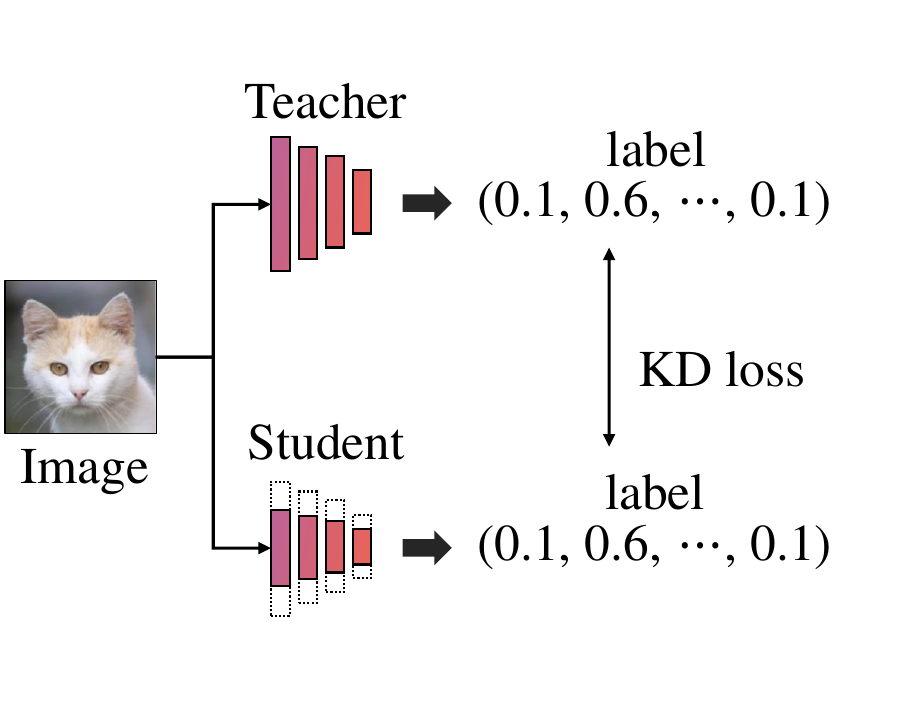}
    \caption{KD for classifier}
    \label{fig:CLSKD}
  \end{subfigure}
  \hfill
  \begin{subfigure}{0.32\linewidth}
    \includegraphics[height=3.3cm]{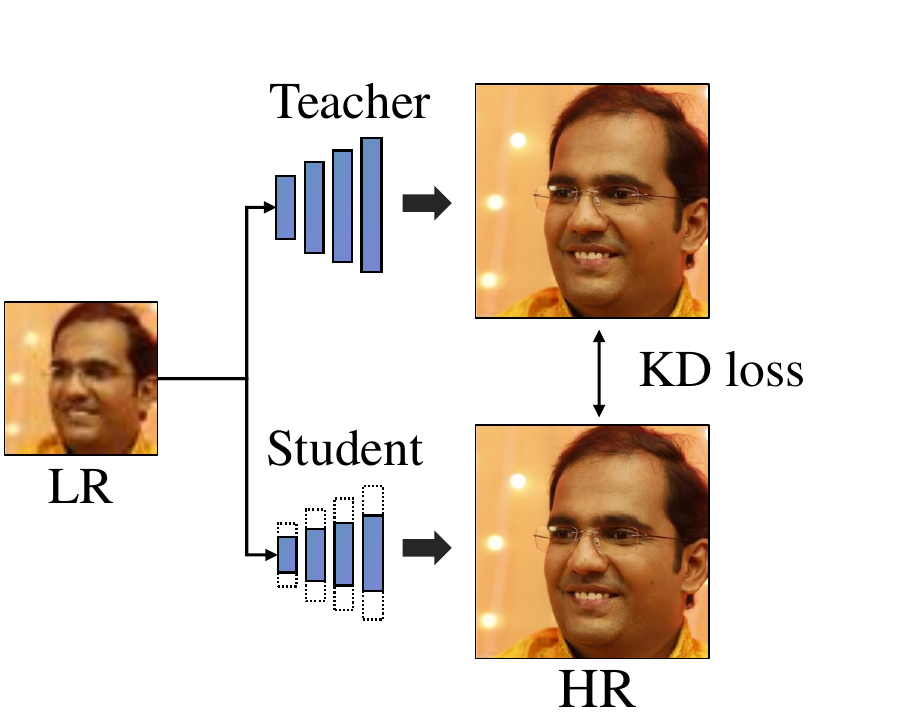}
    \caption{KD for conditional GAN}
    \label{fig:CGKD}
  \end{subfigure}
  \hfill
  \begin{subfigure}{0.32\linewidth}
    \includegraphics[height=3.3cm]{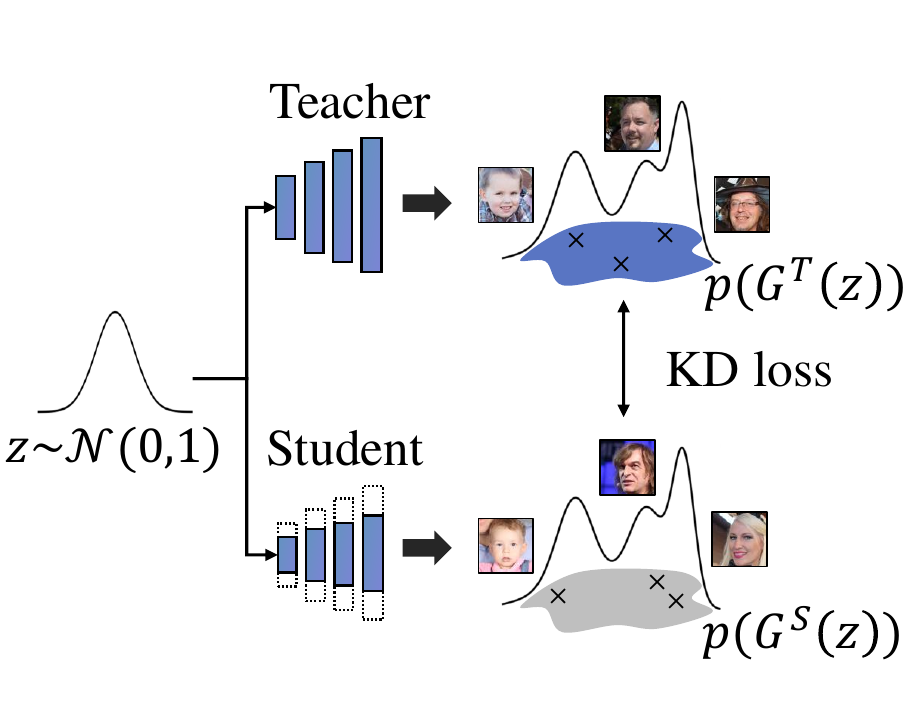}
    \caption{KD for unconditional GAN}
    \label{fig:UCGKD}
  \end{subfigure}  
  \caption{Comparison of knowledge distillation methods. (a) In classification tasks, the instance matching of output labels between the teacher and student is performed. Output labels are in low-dimensional space. Ideally, the outputs of the student and teacher are the same. (b) In conditional generative tasks, the instance matching of output images between the teacher and student is performed. Output images are in high dimensional space. The outputs of the student and teacher are similar (in terms of structure or background). (c) In unconditional generative tasks, the distribution matching of output images between the teacher and student is performed. There is no necessity for each input to have the same output.}
  \label{fig:KD}
\end{figure}

Although compression methods have been extensively studied for classification tasks \cite{lecun1989optimal, hassibi1992second, han2015learning, park2019relational, sreenivasan2022rare}, their na{\"i}ve application to generative models often leads to significant performance degradation \cite{shu2019co, wang2020gan}. As shown in \cref{fig:KD}, to distill the rich knowledge from the teacher to the student, simple label matching is performed in classification models, whereas high-dimensional output matching is required in generative models. Moreover, in GANs, achieving optimal performance requires a delicate balance between the generator and discriminator during adversarial training, which becomes more difficult between the pruned generator and discriminator (see \cref{fig:diseq}).

Recently, several GAN compression methods \cite{li2020gan, liu2021content, wang2020gan, kang2022information, hou2021slimmable, hu2023discriminator, xu2022mind, zhang2022wavelet, li2021revisiting, yeo2023can} have been proposed, but compressing unconditional GAN remains challenging. This is because conditional GANs require instance matching \cite{xu2022mind} as the teacher and student strive for similar outputs in a manner akin to classification tasks, whereas unconditional GAN compression demands distribution matching \cite{kang2022information} (\cref{fig:KD}).
There exist a few unconditional GAN compression studies \cite{wang2020gan, liu2021content, kang2022information, xu2022mind}, but they either still suffer from significant performance degradation \cite{wang2020gan, liu2021content, kang2022information, xu2022mind} or require additional costs such as manual labeling \cite{liu2021content} and MCMC sampling \cite{kang2022information}.

To address these problems, we first propose the Distribution Matching for Efficient compression (\texttt{DiME}). Most GAN compression methods utilized the embedding space (\eg, perceptual \cite{liu2021content, hu2023discriminator}, frequency \cite{zhang2022wavelet}) because directly matching high-dimensional output images leads to significant performance degradation. Similarly, we leverage the foundation models (\ie, DINO \cite{caron2021emerging, oquab2023dinov2}, CLIP \cite{radford2021learning}) as embedding kernels, which have shown successful applications with strong embedding power on various tasks \cite{lin2023clip, zhang2023prompt, kwon2023one}. Furthermore, Santos \etal and Yeo \etal \cite{santos2019learning, yeo2023can} have shown that neural networks can be considered as characteristic kernels to map into Reproducing Kernel Hilbert Space (RKHS), where matching the extracted features of two distributions is equivalent to matching the original distributions as the maximum mean discrepancy (MMD) critic \cite{gretton2006kernel, gretton2012kernel, li2015generative, li2017mmd, santos2019learning}. Additionally, we propose to utilize the global features of the teacher generator to reduce the sampling error. While we ideally hope for the matching of population distributions between the teacher generator ($G^T$) and the student generator ($G^S$) through knowledge distillation, in reality, there is a sampling error due to the matching between sample distributions. Since the distribution of the $G^T$ is fixed, according to the law of large numbers, we can obtain nearly error-free statistics by precomputing a large number of samples from $G^T$. We provide detailed discussion in \cref{sec:ablation}.

\begin{figure}[tb]
  \centering
  \includegraphics[height=6.0cm]{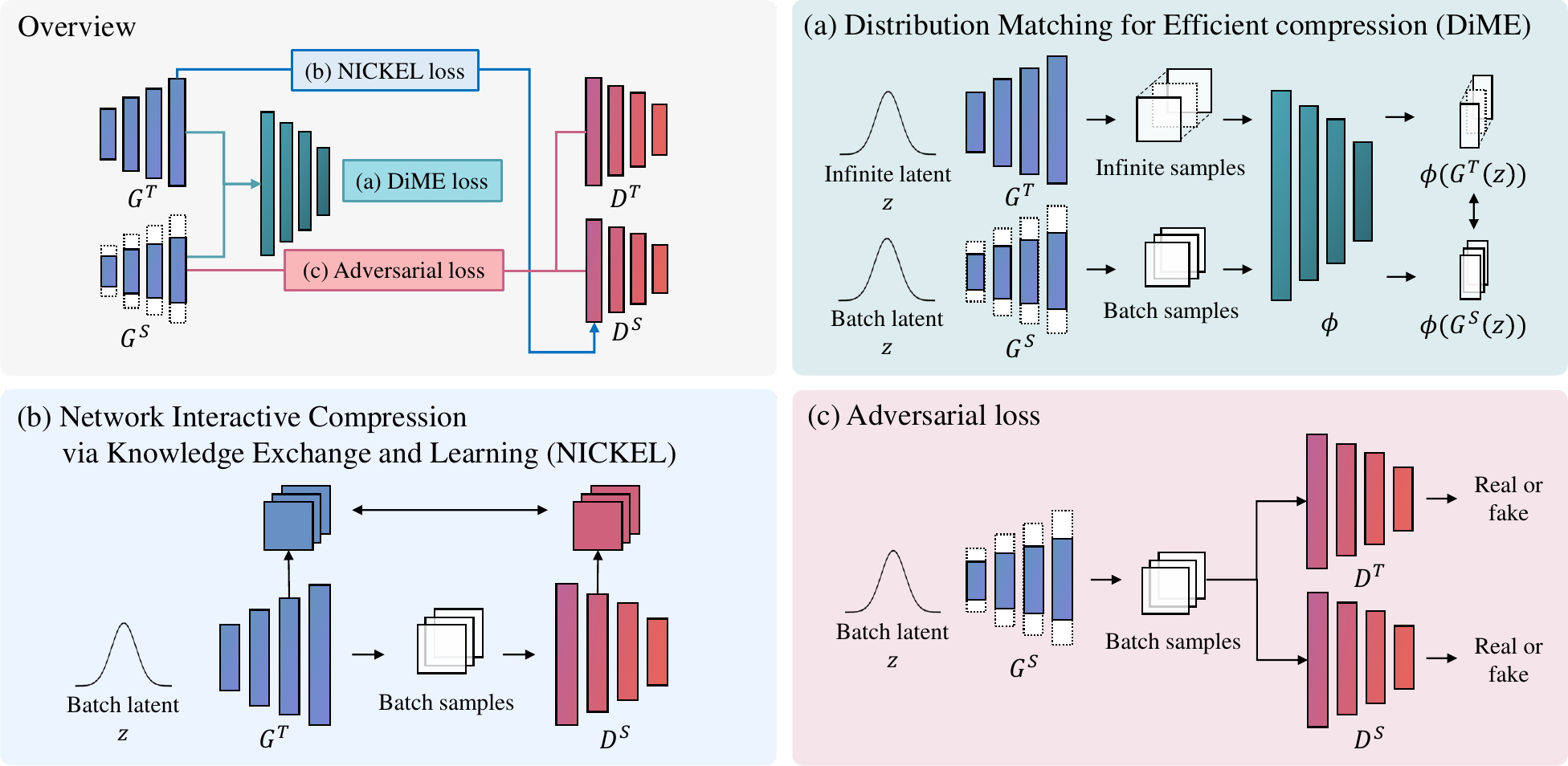}
  \caption{A schematic overview of our method. Our method consists of (a) Distribution Matching for Efficient compression (\texttt{DiME}), (b) Network Interactive Compression via Knowledge Exchange and Learning (\texttt{NICKEL}), and (c) adversarial loss. (a) matches the outputs between the teacher generator ($G^T$) and the student generator ($G^S$) via the foundation model $\phi$ in the embedding space. (b) matches the intermediate features between the teacher generator and the student discriminator ($D^S$). (c) represents the adversarial loss between the student generator and both the teacher discriminator ($D^T$) and the student discriminator.
  }
  \label{fig:example}
\end{figure}

In addition to \texttt{DiME}, to exploit the characteristic of GAN that consists of a generator and a discriminator, we propose Network Interactive Compression via Knowledge Exchange and Learning (\texttt{NICKEL}). In GAN training, Lee \etal \cite{lee2022generator} has shown that the discriminator can provide more meaningful signals as feedback by learning the semantic knowledge of the generator. Inspired by Lee \etal, we not only distill knowledge directly between the generators (\ie., \texttt{DiME}), but also distill knowledge from the more informative $G^T$ to the student discriminator ($D^S$) by transmitting knowledge between generators via the discriminator indirectly. By utilizing $G^T$, we obtain two distinct advantages, Firstly, from the onset of training, $D^S$ learns the rich semantic knowledge embedded within the $G^T$. Secondly, the $G^T$ provides a wealth of knowledge surpassing that of $G^S$. Furthermore, we observe that \texttt{NICKEL} enhances the stability of GAN compression (see \cref{fig:stability}). To the best of our knowledge, \texttt{NICKEL} is the first method that distills the knowledge from $G^T$ to $G^S$ via the feedback of $D^S$ for model compression.

Our experimental results show that \texttt{DiME} outperforms existing state-of-the-art compression methods through knowledge distillation between $G^T$ and $G^S$. 
By applying \texttt{DiME} to StyleGAN2, which has a baseline FID of 4.02, resulted in FID scores of 11.25 and 18.32 at compression rates of 95.87\% and 98.92\%, respectively. This compares favorably to the state-of-the-art method~\cite{kang2022information}, which achieves FID scores of 14.01 and 22.23 at the same compression rate. This demonstrates the power of using foundation models as embedding kernels for knowledge distillation. In addition, by using \texttt{NICKEL} with \texttt{DiME}, we further enhance the FID scores to 10.45 and 15.93 with improved stability, setting a new standard in GAN compression performance. It is worth to note that we achieve a reasonable performance with the FID of 29.38 at the extreme compression rate of 99.69\%, surpassing the previous state-of-the-art performance by a significant margin.
Our contributions can be summarized as follows:

\begin{figure}[tb!]
  \centering
  \begin{subfigure}{0.324\linewidth}
    \includegraphics[height=2.85cm]{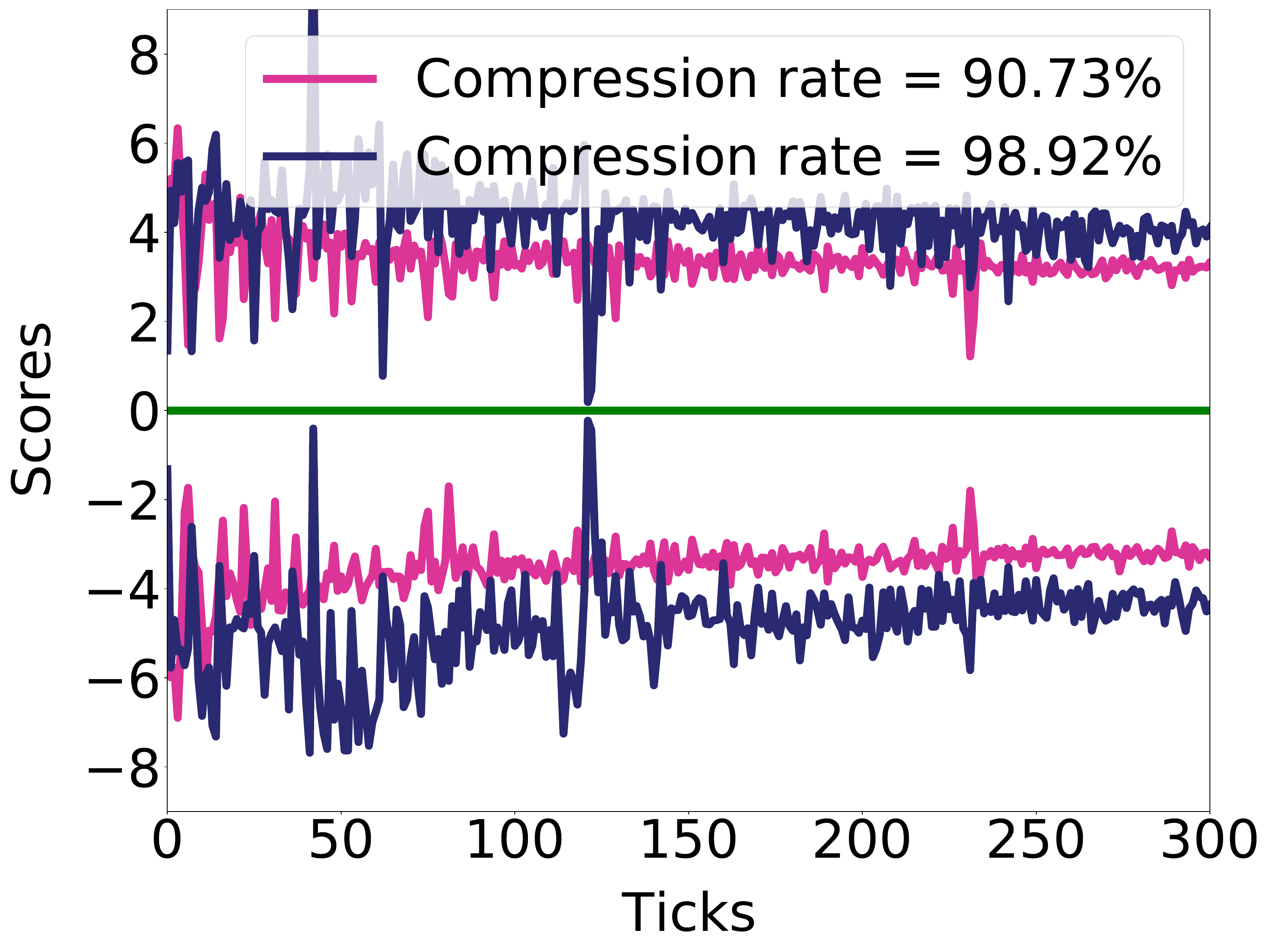}
    \caption{Logits of D on ITGC\cite{kang2022information}}
    \label{fig:diseq}
  \end{subfigure}
  \hfill
  \begin{subfigure}{0.324\linewidth}
    \includegraphics[height=2.85cm]{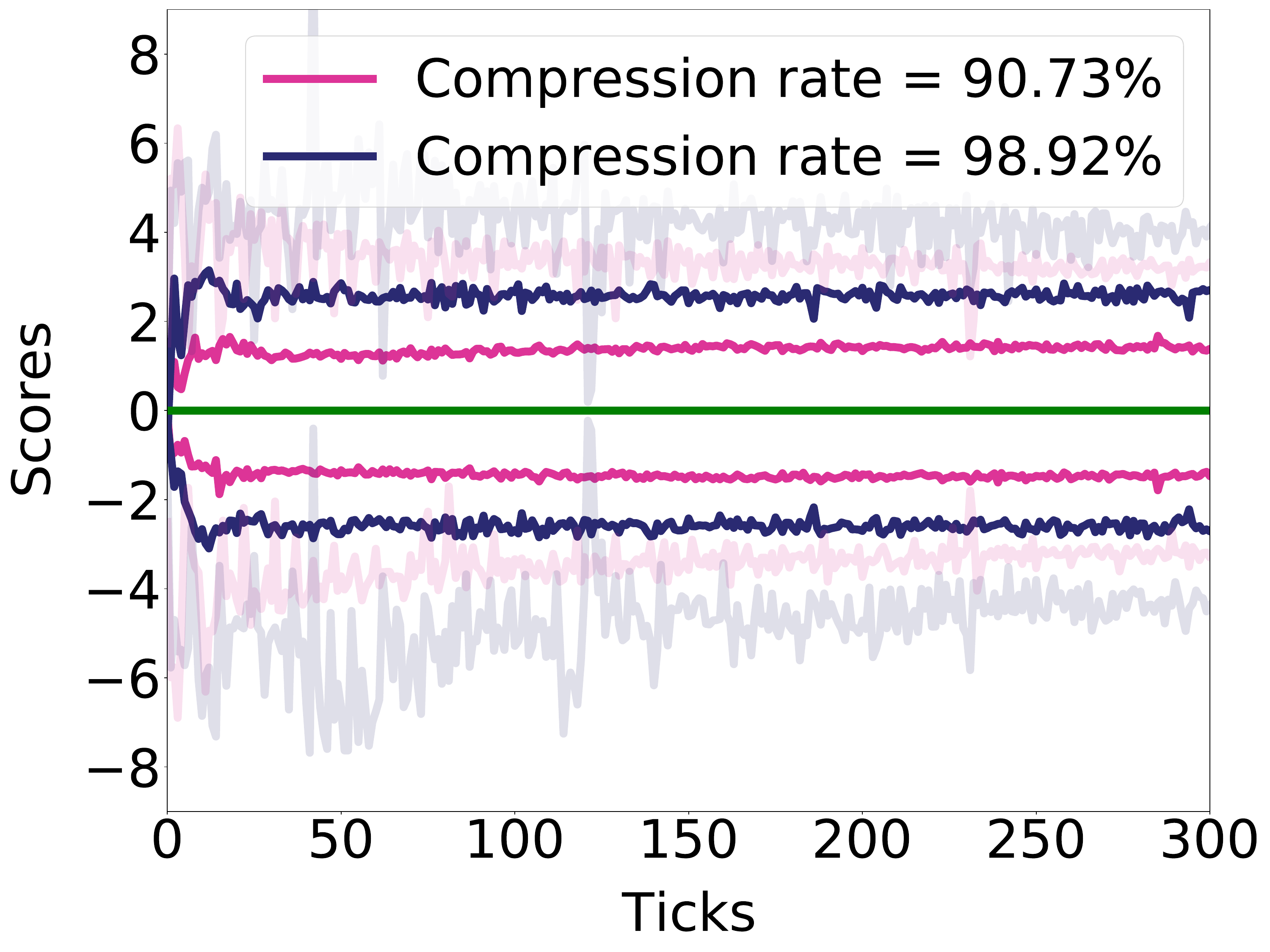}
    \caption{Logits of D on \texttt{NICKEL \& DiME}}
    \label{fig:eq}
  \end{subfigure}
  \hfill
  \begin{subfigure}{0.324\linewidth}
    \includegraphics[height=2.85cm]{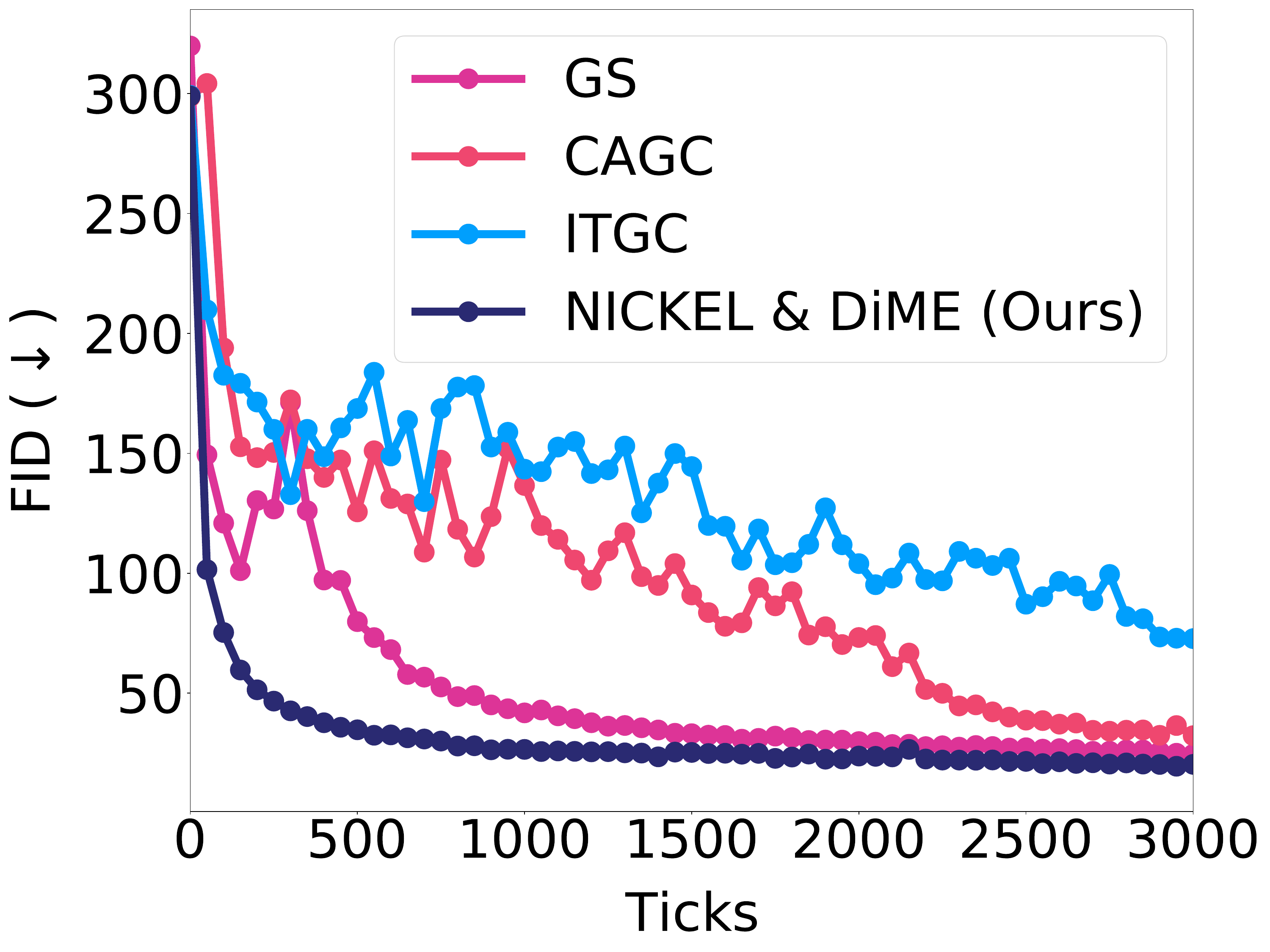}
    \caption{Convergence of FID}
    \label{fig:convergence}
  \end{subfigure}  
  \caption{Comparison of stability of ours and state-of-the-art compression methods. (a) indicates the logits of the discriminator for the pruned generator on ITGC\cite{kang2022information}. The green solid line represents the ideal equilibrium state. When the compression rate is 98.92$\%$ (blue dash line), it shows a more severe imbalance state compared to when the compression rate is 90.73$\%$ (red dash line). (b) indicates the logits of the discriminator for the pruned generator on \texttt{NICKEL \& DiME}. Our method mitigates the imbalance between the discriminator and the pruned generator. (c) indicates the FID convergence plot when the compression rate is 98.92\%. \texttt{NICKEL \& DiME} converges the most stably.}
  \label{fig:stability}
\end{figure}

\begin{itemize}
    \item We propose \texttt{DiME}, an effective distillation method for GANs that ensures the matching output distributions between $G^T$ and $G^S$ via employing foundation models as kernels for MMD loss (\cref{section:DiME}). \texttt{DiME} outperforms existing GAN compression methods, achieving state-of-the-art performance in GAN compression at all compression rates.
    \item We propose \texttt{NICKEL} that further enhances the distillation capability by providing more meaningful feedback from $D^S$. We observe that \texttt{NICKEL} leads to the improvement of stability (\cref{section:NICKEL}).
    \item With \texttt{NICKEL \& DiME}, our final model further raises the bar of the state-of-the-art. Our method shows a stable convergence with competitive performance, even at the extremely high compression rates of 99.69\% (\cref{section:exp}).
    \item Last but not least, we standardize and benchmark GAN compression methods using official codes, ensuring future compatibility and reproducibility.
\end{itemize}

\section{Related Work}
\label{section:rw}

\subsection{GAN Compression}
Most GAN compression methods have been explored in conditional GAN settings \cite{li2020gan, li2021revisiting, zhang2022wavelet, hu2023discriminator}, which are unsuitable for distribution matching in unconditional GAN compression (see \cref{fig:KD}). Occasionally, to address this problem, GAN compression methods have been explored \cite{liu2021content, kang2022information, xu2022mind}, proposing better embedding spaces or distance metrics between $G^T$ and $G^S$. Wang \etal \cite{wang2020gan} proposed the GAN slimming (GS), a unified optimization framework, and emphasized that na{\"i}ve application of compression methods leads to significant performance degradation due to the notorious instability of GANs. Liu \etal \cite{liu2021content} proposed the content-aware GAN compression (CAGC) method, which focuses on only the contents of interest (\eg, object, face) to distill the knowledge, but this method requires additional costs due to manual labeling of contents. Li \etal \cite{li2021revisiting} proposed the generator-discriminator cooperative compression (GCC) to maintain the nash-equilibrium between $G^S$ and $D^S$, but nash-equilibrium still cannot be maintained, in complex settings. Kang \etal \cite{kang2022information} proposed the information-theoretic GAN compression (ITGC) by maximizing the mutual information between $G^T$ and $G^S$. ITGC requires a lot of computational costs due to the energy-based model and MCMC sampling. Xu \etal \cite{xu2022mind} proposed the StyleKD that focuses on the mapping network to achieve consistent outputs between $G^T$ and $G^S$. However, StyleKD can only be applied to networks based on StyleGAN. To address these issues, we propose Distribution Matching for Efficient compression (\texttt{DiME}), which matches the distribution between $G^T$ and $G^S$ via foundation kernels.

\subsection{Discriminator Regularization}
Generally, GAN compression methods are focused on the $G^S$, thus it is applied in the form of generator regularization for knowledge distillation. On the other hand, GCC \cite{li2021revisiting} emphasized the importance of considering not only the generator but also the discriminator to maintain the Nash equilibrium state between the compressed generator and discriminator. Similar phenomena were observed by several studies \cite{chen2021gans, zhang2022wavelet, hu2023discriminator}. To address this issue, GCC used the selective activation discriminator, which partially activates the channels of the discriminator by utilizing the capacity constraint to maintain the Nash equilibrium state. However, GCC still shows significant performance degradation due to instability. In GAN training, Lee \etal \cite{lee2022generator} proposed generator-guided discriminator regularization (GGDR). GGDR showed that the discriminator can learn the semantic knowledge from the generator and lead to performance improvement of the generator by providing more powerful adversarial loss as feedback. However, GGDR cannot inject meaningful knowledge of the generator into the discriminator in the early stage because the initial generator is close to being a randomly initialized generator. Inspired by GGDR, we propose the Network Interactive Compression via Knowledge Exchange and Learning (\texttt{NICKEL}), which distills the knowledge from $G^T$ to $D^S$ and encourages powerful feedback from $D^S$ to $G^S$.

\section{Method}
\label{section:method}
\subsection{Knowledge Distillation with Foundation Kernels MMD}
\label{section:DiME}

Generally, knowledge distillation (KD) minimizes the distance $d_{kd}$ (\eg, wavelet loss \cite{zhang2022wavelet}) between the outputs of $G^T$ and $G^S$, encouraging $G^S$ to mimic $G^T$. We can achieve more effective knowledge distillation by designing a better distance metric. In this paper, we propose Distribution Matching for Efficient compression (\texttt{DiME}), which matches the distributions between $G^T$ and $G^S$ in the space embedded by foundation kernels $\phi$ as distance $d_{kd}$:
\begin{equation}
\mathcal{L}_{KD} = d_{kd}(G^T(z), G^S(z)) = \mathbb{E}[||\phi(G^T(z)), \phi(G^S(z))||_1]
\label{eq:foundation_KD}
\end{equation}
This is equivalent to using the MMD critic \cite{gretton2006kernel, gretton2012kernel, li2015generative, li2017mmd, santos2019learning}, a statistical method that matches two distributions in RKHS, assuming that the foundation kernels $\phi$ are characteristic kernels \cite{santos2019learning, yeo2023can}.

While \cref{eq:foundation_KD} generally shows good performance, we observe the tremendous performance degradation of all baselines (\ie, GS, CAGC, ITGC, \cref{eq:foundation_KD}) when $G^S$ has extremely few parameters (see \cref{fig:convergence}). As shown in \cref{fig:diseq}, the Nash equilibrium breaks down when $G^S$ has fewer parameters, which consequently leads to the performance degradation of adversarial loss. 
To improve the stability of KD loss
in the early stage, we utilize the global features of $G^T$. The global features are computed by inferring over a multitude of images rather than batch images, enabling the calculation of popular distribution statistics. Utilizing the global features mitigates the sampling error induced by the batch size in KD, with detailed discussion included in \cref{sec:ablation}.

\subsection{Network Interactive Compression via Knowledge Exchange and Learning}
\label{section:NICKEL}
GAN utilizes a discriminator, which is a learnable network as the loss during the training of the generator. The performance of the generator is heavily influenced by the quality of feedback provided by the discriminator. GGDR \cite{lee2022generator} showed that during GAN training, the discriminator can learn semantic knowledge from the generator. Subsequently, the discriminator provides better feedback to the generator, thus improving the performance of the generator. Inspired by GGDR, we propose \texttt{NICKEL}, which distills knowledge from $G^T$ into $D^S$ to provide more powerful feedback to $G^S$. \texttt{NICKEL} has advantages over simply applying GGDR to $G^S$ for two reasons. First, GGDR may struggle to provide meaningful information when $G^S$ resembles a random network during early training, whereas \texttt{NICKEL} can distill rich information from $G^T$ from the outset. Second, in GAN compression, due to the smaller network structure of $G^S$, GGDR cannot provide knowledge as rich as $G^T$. Therefore, we propose fine-tuning $D^S$ via \texttt{NICKEL} to learn information from $G^T$. However, fine-tuning $D^S$ using the \texttt{NICKEL} loss alone is insufficient to fully leverage the information from the pre-trained discriminator. Therefore, for adversarial learning, both $D^T$ and $D^S$ are employed. The loss function of \texttt{NICKEL} can be formulated as follows:
\begin{align}
& \mathcal{L}_{\texttt{NICKEL}} = \sum_{i=1}^L d_{\texttt{NICKEL}}(G_i^T(z), f_i(D_i^S(G^T(z)))),
\label{NICKEL}
\end{align}
where $D_i^S(G^T(z))$ and $G_i^T(z)$ represent the feature maps of the $i$-th layer of $D^S$ and $G^T$, respectively. $f_i$ is a linear transform to match the shape of feature maps. As Lee \etal \cite{lee2022generator} mentioned, the knowledge of the generator contains a lot of semantic information. Therefore, we utilize the wavelet loss \cite{zhang2022wavelet} for $d_{\texttt{NICKEL}}$, which is good for matching semantic information.

\subsection{Training Objective}
In summary, our training loss for GAN compression is formulated as:
\begin{equation}
\mathcal{L} = \mathcal{L}_{adv} + \lambda_{dino} \cdot \mathcal{L}_{dino} + \lambda_{clip} \cdot \mathcal{L}_{clip} + \lambda_{\texttt{NICKEL}} \cdot \mathcal{L}_{\texttt{NICKEL}},
\label{training_obj}
\end{equation}
where $\lambda_{dino}$ and $\lambda_{clip}$ are the weights for the knowledge distillation, which utilizes the dino embedding and clip embedding, respectively. $\lambda_{\texttt{NICKEL}}$ is the weight for \texttt{NICKEL} loss $\mathcal{L}_{\texttt{NICKEL}}$ in \cref{NICKEL}. $\mathcal{L}_{adv}$ is the adversarial loss, which is the min-max objective function that includes both $D^T$ and $D^S$. $\mathcal{L}_{dino}$ and $\mathcal{L}_{clip}$ are the knowledge distillation losses of \texttt{DiME} in \cref{eq:foundation_KD}. 

\section{Experiments}
\label{section:exp}
\subsection{Setups}

\subsubsection{Implementation details.} We set 20, 15, and 10 for $\lambda_{dino}$, $\lambda_{clip}$, and $\lambda_{\texttt{NICKEL}}$, respectively. We use the same pruned generator as CAGC, and for CLIP and DINO, we use the pretrained weights.\footnote{\href{https://openaipublic.azureedge.net/clip/models/afeb0e10f9e5a86da6080e35cf09123aca3b358a0c3e3b6c78a7b63bc04b6762/RN50.pt}{CLIP.pt} and \href{https://github.com/OpenGVLab/CaFo?tab=readme-ov-file}{DINO.pt}} For \texttt{NICKEL}, we match feature maps in 1/4 resolution channels of the generator's output (e.g., 64x64 feature maps for 256 resolution) on LH, HL, and HH components of Haar wavelet. To obtain the global features, we conduct 20,000 model inferences with the batchsize of 256.

\begin{table}[t!]
\caption{Comparison of FID scores of our methods (\texttt{DiME} and \texttt{NICKEL \& DiME}) and state-of-the-art compression methods on StyleGAN2 for various datasets. The dagger symbol indicates the performance of the official model provided on the original paper's GitHub repository. The right side of the arrow shows the results obtained using NVIDIA's official FID measurement code. Our reproduced models achieve higher performance compared to the official models.}

\centering
\resizebox{\textwidth}{!}{
    \begin{tabular}{ccccccc}
    \toprule
    \multirow{2}{*}{Model}    & \multirow{2}{*}{Dataset}     & \multirow{2}{*}{Method} & \multirow{2}{*}{\#params.} & \multirow{2}{*}{FLOPs} & \multirow{2}{*}{\shortstack{Compression\\rate}} & \multirow{2}{*}{{FID$\downarrow$}} \\ \\ \midrule
    \multirow{28}{*}{StyleGAN2} & \multirow{19}{*}{\shortstack{FFHQ\\(256$\times$256)}}     & (1) Full model \cite{karras2020training}       &  24.77M       & 14.90B           & -    & 4.02        \\ 
                                &                                                      & (2) Full model\textdagger & 30.03M & 45.12B & -& 4.5 \\    \cmidrule(l){3-7}
                              &                              & (3) CAGC\textdagger \cite{liu2021content}    & \multirow{2}{*}{5.57M}       & \multirow{2}{*}{4.12B}  & \multirow{2}{*}{90.87\%}     & 7.9 $\rightarrow{}$10.82    \\
                              &                              & (4) DCP-GAN\textdagger \cite{chung2024diversity}                       &                              &                         &                        & 6.35 $\rightarrow{}$8.93        \\         \cmidrule(l){3-7}   &                              & CAGC                                          & \multirow{4}{*}{8.72M}       & \multirow{4}{*}{3.73B}  & \multirow{4}{*}{74.96\%}      & 5.24        \\
                              &                              & ITGC \cite{kang2022information}               &        &   &      & 5.27        \\
                              &                              & \texttt{DiME}                                 &        &                         &                               & 5.00        \\
                              &                              & \texttt{NICKEL \& DiME}                       &            &                         &                               & \textbf{4.42}        \\
                              \cmidrule(l){3-7}                              
                              &                              & GS \cite{wang2020gan}                         & \multirow{6}{*}{4.96M}       & \multirow{6}{*}{1.38B}  & \multirow{6}{*}{90.73\%}      & 10.26        \\
                              &                              & CAGC                     &        &                         &                               & 10.06        \\
                              &                              & GCC \cite{li2021revisiting}                   &        &                         &                               & 11.19        \\
                              &                              & ITGC                                          &        &                         &                               & 10.02        \\
                              &                              & \texttt{DiME}                                 &        &                         &                               & 8.39        \\
                              &                              & \texttt{NICKEL \& DiME}                       &        &                         &                               & \textbf{7.43}        \\
                              \cmidrule(l){3-7}
                              &                              & CAGC                                          & \multirow{4}{*}{2.69M}       & \multirow{4}{*}{0.16B}  & \multirow{4}{*}{98.92\%}      & 23.05        \\
                              &                              & ITGC                                          &        &                         &                               & 22.23        \\
                              &                              & \texttt{DiME}                                 &        &                         &                               & 18.32        \\
                              &                              & \texttt{NICKEL \& DiME}                       &        &                         &                               & \textbf{15.93}        \\
                              \cmidrule(l){2-7}
                              & \multirow{5}{*}{\shortstack{FFHQ\\(1024$\times$1024)}}    & Full model           & 30.37M  & 74.27B                   & -         &   2.74          \\ 
                              &                                                      & Full model\textdagger & 49.1M & 74.3B & -& 2.7 \\    \cmidrule(l){3-7}\cmidrule(l){3-7}
                              &                              & (5) CAGC\textdagger                                            & \multirow{2}{*}{9.2M}  & \multirow{2}{*}{7.0B}   & \multirow{2}{*}{89.39\%}   &   7.6 $\rightarrow{}$7.53          \\
                              &                              & (6) DCP-GAN\textdagger                                          &   &                          &    & 5.80 $\rightarrow{}$5.87                   \\
                              &                              & \texttt{NICKEL \& DiME}                       &  5.65M     &  6.99B                        & 90.59\%   & \textbf{6.41}               \\ 
                              \cmidrule(l){2-7}                              
                              & \multirow{6}{*}{\shortstack{LSUN\\Church\\(256$\times$256)}}    & Full model           & 30.03M  & 45.12B                   & -         &   3.97          \\    \cmidrule(l){3-7}
                              &                              & CAGC                                            & \multirow{4}{*}{5.57M}  & \multirow{4}{*}{4.12B}   & \multirow{4}{*}{90.87\%}   &   4.50          \\
                              &                              & StyleKD                                          &   &                          &    & 4.47                   \\
                              &                              & (7) DCP-GAN\textdagger                                          &   &                          &    & 4.87 $\rightarrow{}$4.82                   \\
                              &                              & \texttt{NICKEL \& DiME}                       &       &                          &    & \textbf{3.94}               \\    
                                                                        
                              \cmidrule(l){1-7} \multirow{2.5}{*}{DDPM}    & \multirow{2.7}{*}{\shortstack{LSUN\\Church\\(256$\times$256)}}    & Full model\textdagger    & 113.7M             & 497.4B       & -          & 10.6        \\ \cmidrule(l){3-7}
                              &                          & SPDM (t=100)\textdagger \cite{fang2024structural}    & 63.2M             & 277.6B       & 44.19\%          & 13.9        \\              
                              \bottomrule
    \end{tabular}
}
\label{table:quan_stylegan}
\end{table}

\begin{table}[t!]
\caption{Comparison of FID scores of our methods (\texttt{DiME} and \texttt{NICKEL \& DiME}) and state-of-the-art compression methods on various architectures for various datasets.}
\centering
\resizebox{\textwidth}{!}{
    \begin{tabular}{ccccccc}
    \toprule
    \multirow{1}{*}{Model}    & \multirow{1}{*}{Dataset}     & \multirow{1}{*}{Method} & \#params. &\multirow{1}{*}{FLOPs} & Compression rate & \multirow{1}{*}{FID$\downarrow$} \\  \midrule
    \multirow{9}{*}{SNGAN}    & \multirow{9}{*}{\shortstack{CIFAR-10\\(32$\times$32)}}    & Full model \cite{miyato2018spectral}   & 4.28M             & 3.36B       & -          & 17.71        \\ \cmidrule(l){3-7}
                              &                              & CAGC \cite{liu2021content}            & \multirow{4}{*}{1.20M} & \multirow{4}{*}{0.85B}         & \multirow{4}{*}{74.88\%}       & 40.45        \\
                              &                              & ITGC \cite{kang2022information}        &            &                    &                              & 43.66       \\
                              &                              & \texttt{DiME}                                   &    &                            &                              & 31.98        \\
                              &                              & \texttt{NICKEL \& DiME}                    &          &                      &                              & \textbf{23.11}        \\ \cmidrule(l){3-7}
                              &                              & CAGC             & \multirow{4}{*}{0.50M} & \multirow{4}{*}{0.31B}         & \multirow{4}{*}{90.85\%}       & 51.93        \\
                              &                              & ITGC        & &                                &                              & 59.99
        \\
                              &                              & \texttt{DiME}                          &         &                                &                              & 36.89        \\
                              &                              & \texttt{NICKEL \& DiME}                &    &                                &                              & \textbf{28.27}        \\ \cmidrule(l){1-7}
                              \multirow{3.5}{*}{BigGAN}    & \multirow{3.5}{*}{\shortstack{CIFAR-10\\(32$\times$32)}}    & Full model \cite{brock2018large}    & 8.83M             & 3.83B       & -          & 10.66        \\ \cmidrule(l){3-7}
                              &                          & CAGC    & \multirow{3}{*}{0.89M}             & \multirow{3}{*}{0.37B}       & \multirow{3}{*}{90.45\%}          & \textbf{26.32}        \\
                              &                          & ITGC    &              &        &           & 27.78        \\
                                                         &                          & \texttt{NICKEL \& DiME}    &              &        &           & 26.66        \\ \cmidrule(l){1-7}
                              \multirow{13}{*}{StyleGAN2}    & \multirow{4.5}{*}{\shortstack{CelebA\\(128$\times$128)}}    & Full model \cite{karras2020training}    & 24.53M             & 11.27B       & -          & 2.70        \\ \cmidrule(l){3-7}
                              &                          & CAGC    & \multirow{4}{*}{3.08M}             & \multirow{4}{*}{0.27B}       & \multirow{4}{*}{97.60\%}          & 10.05        \\
                              &                          & ITGC    &              &        &           & 10.09        \\
                              &                          & StyleKD    &              &        &           & 10.41        \\
                             &                          & \texttt{NICKEL \& DiME}    &              &        &           & \textbf{7.56}        \\ \cmidrule(l){2-7}
                                                        & \multirow{2.5}{*}{\shortstack{AFHQ\\(512$\times$512)}}    & Full model    & 30.28M             & 59.67B       & -          & 3.01        \\ \cmidrule(l){3-7}
                             &                          & \texttt{NICKEL \& DiME}    & 10.30M             & 14.94B       & 74.96\%          & \textbf{3.17}        \\ \cmidrule(l){2-7}
                              & \multirow{6.5}{*}{\shortstack{LSUN\\CAT\\(256$\times$256)}}    & Full model           & 24.77M  & 14.90B                   & -         &   8.19          \\    \cmidrule(l){3-7}
                              &                              & GS                                            & \multirow{5}{*}{4.96M}  & \multirow{5}{*}{1.38B}   & \multirow{5}{*}{90.73\%}   &   17.11          \\
                              &                              & CAGC                                          &   &                          &    & 12.31                   \\
                              &                              & ITGC                                          &   &                          &    & 12.06                    \\
                              &                              & \texttt{DiME}                                 &            &                          &    & 11.59                  \\
                              &                              & \texttt{NICKEL \& DiME}                       &       &                          &    & \textbf{10.80}               \\ 
                              
                              \bottomrule
    \end{tabular}
}
\label{table:quan_sngan}
\end{table}

\begin{figure}[tb!]
  \centering
  \begin{subfigure}{0.3\linewidth}
    \includegraphics[height=2.85cm]{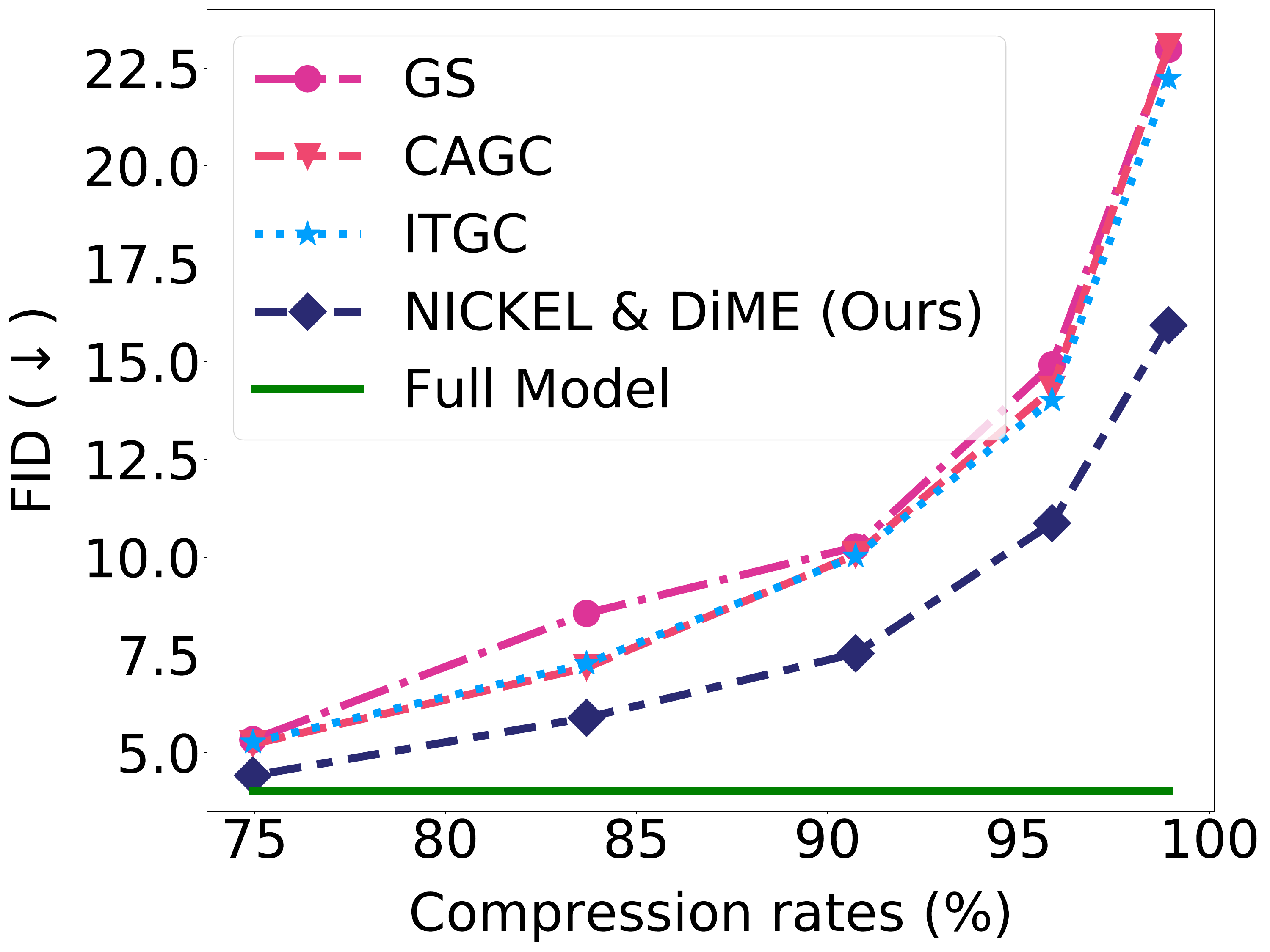}
    \caption{Comparison of FID}
    \label{fig:fid}
  \end{subfigure}
  \hfill
  \begin{subfigure}{0.3\linewidth}
    \includegraphics[height=2.85cm]{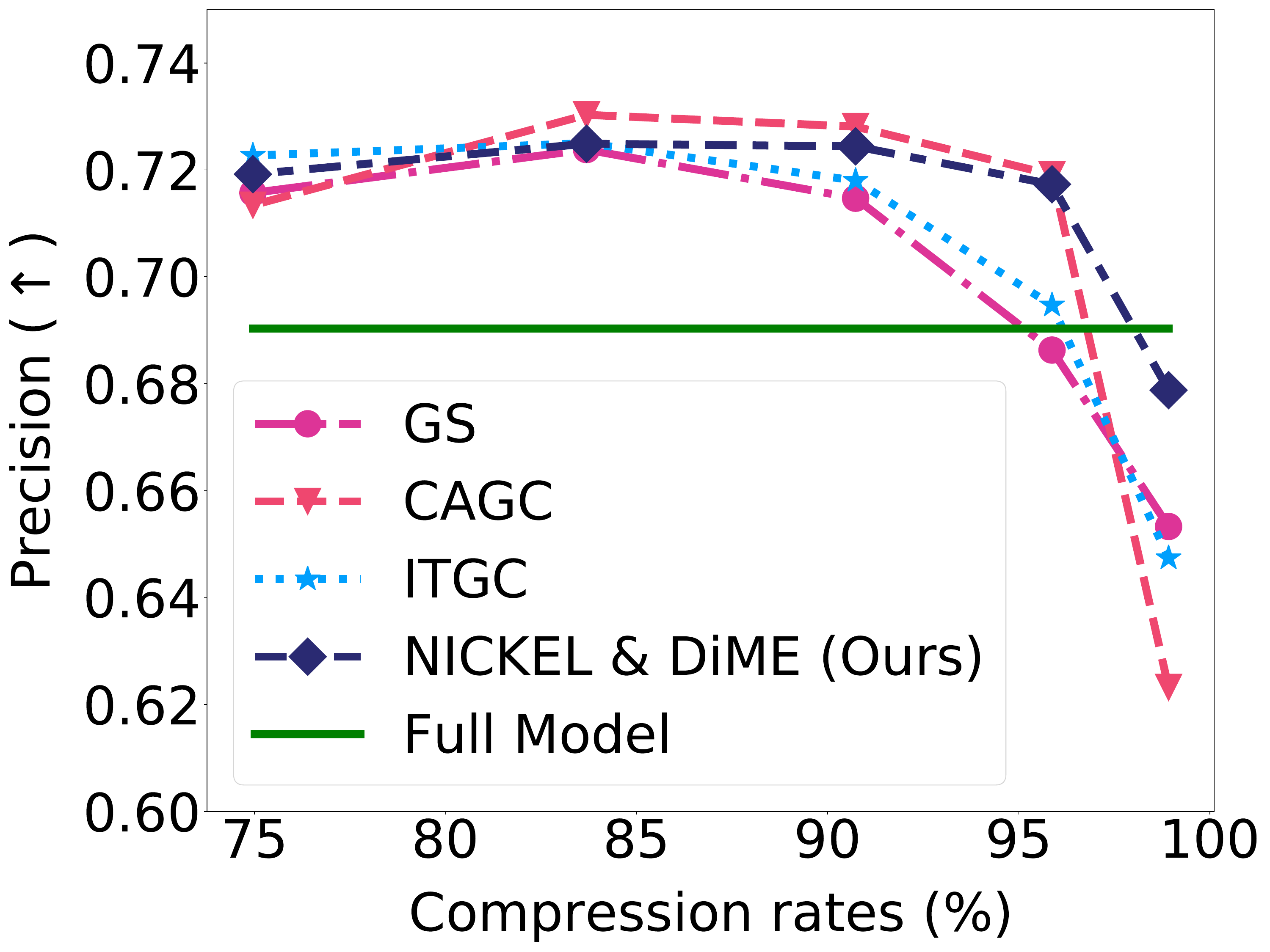}
    \caption{Comparison of Precision}
    \label{fig:precision}
  \end{subfigure}
  \hfill
  \begin{subfigure}{0.3\linewidth}
    \includegraphics[height=2.85cm]{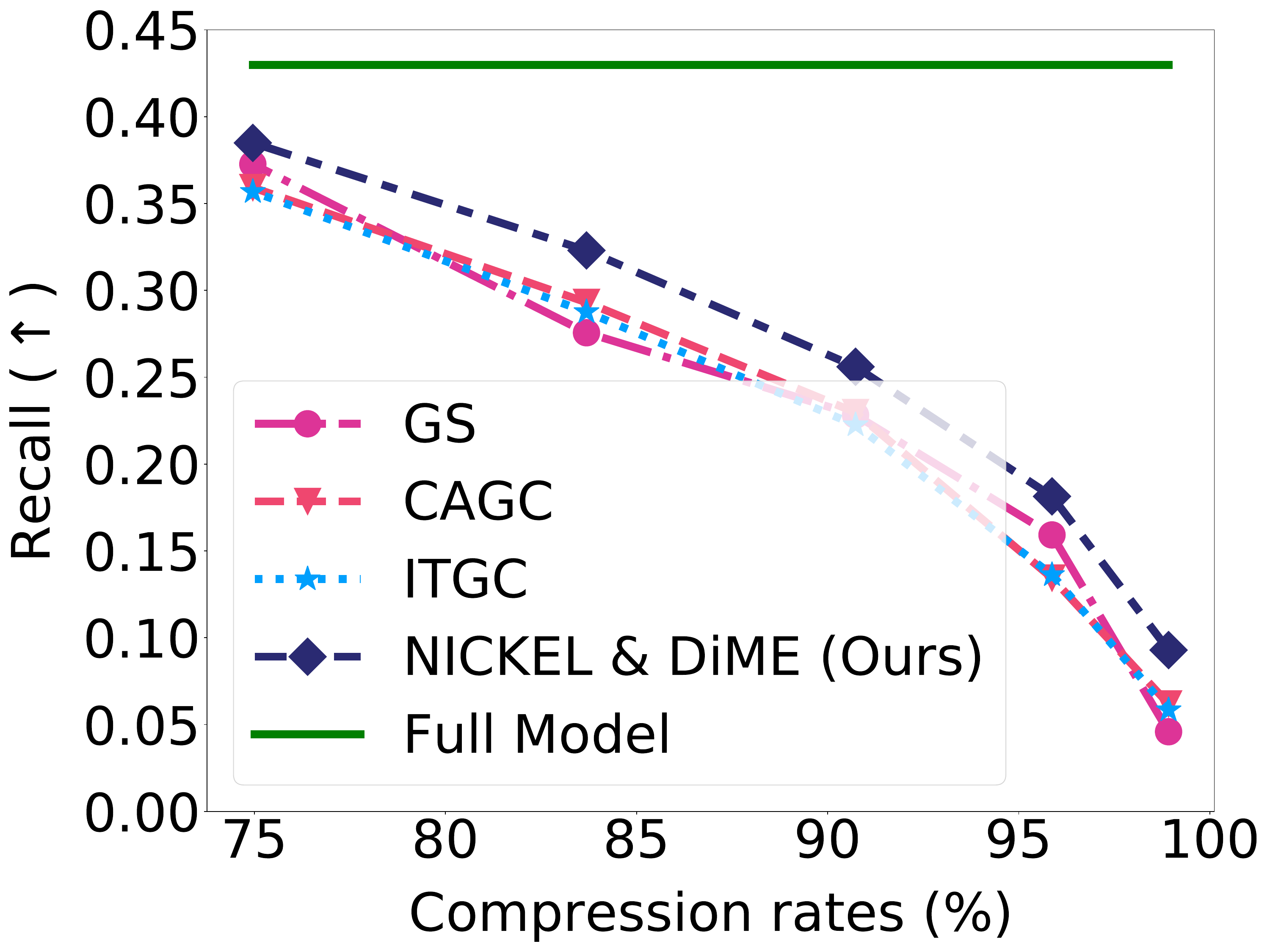}
    \caption{Comparison of Recall}
    \label{fig:recall}
  \end{subfigure}  
  \caption{Performance comparison as a function of compression rates on StyleGAN2 for FFHQ. (a) indicates a function showing how FID varies with compression rates. \texttt{NICKEL \& DiME} consistently outperforms other state-of-the-art compression methods at various compression rates. At a compression rate of 74.96\%, \texttt{NICKEL \& DiME} shows only 9.68\% performance degradation compared to the full model, and the performance degradation due to increasing compression rates occurs less than other state-of-the-art compression methods. (b) indicates a function showing how Precision varies with compression rates. \texttt{NICKEL \& DiME} shows comparable fidelity scores to other methods. (c) indicates a function showing how Recall varies with compression rates. \texttt{NICKEL \& DiME} shows better preservation of diversity compared to other methods, even with higher compression rates. 
}
  \label{fig:performance}
\end{figure}

\subsubsection{Datasets and Evaluation Metrics.}
 We use CelebA \cite{liu2015deep}, FFHQ \cite{karras2019style}, AFHQ \cite{choi2020stargan}, LSUN-Church and LSUN-CAT \cite{yu2015lsun} datasets, and CIFAR-10 \cite{krizhevsky2009learning}. 
se Fr\'echet Inception Distance (FID)~\cite{heusel2017gans} and  Precision \& Recall~\cite{kynkaanniemi2019improved, naeem2020reliable, kim2024topp} to evaluate the performance of the GANs, measuring both the quality and diversity of the generated images. We note the average of the five best FID scores. , we use the \texttt{fvcore} library to measure FLOPs, ensuring consistency with reported FLOPs and enabling fair comparisons with existing works.\footnote{If you noticed any discrepancy in FLOPs, the discrepancy stems from (a) the typo in the original CAGC paper, where FLOPs were reported as 4.1M instead of 4.1B---a mistake recognized in subsequent papers; and (b) different libraries used. The reported FLOPs of 4.1B at 90.87\% compression can be reproduced via \texttt{fvcore} library. In contrast, when used \texttt{torchprofile} library with the official NVIDIA StyleGAN2 code, it gives the FLOPs of 5.33B at 90.73\%.}

\subsubsection{Baselines.}
We follow the architecture and training setups of StyleGAN2~\cite{karras2020training}, except for augmentation (not used). Our pruned generators are identical to CAGC \cite{liu2021content}. The architecture and training setups for BigGAN and SNGAN follow Kang \etal \cite{kang2023studiogan}. We use the official code of StyleGAN2-ADA-PyTorch and StudioGAN. We also compare Structural Pruning for Diffusion Models (SPDM) \cite{fang2024structural}.

\subsection{Benchmarking and Reproducibility in GAN Compression}
The performance metrics for previous GAN compression techniques were often based on outdated and unofficial code. To address this and for future compatibility and reproducibility in the field, we reimplement the state-of-the-art GAN compression methods using NVIDIA's official StyleGAN2-ADA code. The official code shows comparable performance to the unofficial code, but with more efficient FLOPs and fewer parameters (see (1) and (2) in \cref{table:quan_stylegan}). Additionally, we provide results re-evaluated using NVIDIA's official FID code, utilizing trained weights from the original GitHub repository of the baselines (indicated by the right of the arrow of (3)-(7) in \cref{table:quan_stylegan}). While the re-evaluated FID scores for 1024$\times$1024 FFHQ and LSUN-Church datasets are similar to the reported FID, we find that the FID for the 256$\times$256 FFHQ dataset shows a significant performance gap (see (3) and (4) in \cref{table:quan_stylegan}). This discrepancy arises because the reported FID scores were not calculated by computing the feature embedding directly but by using the feature embedding provided by CAGC's custom FID code. We hypothesize that the feature embedding provided by CAGC may be biased for the 256$\times$256 FFHQ dataset.

\subsection{Results}

We first compare the knowledge distillation performance of \texttt{DiME}, as described in \cref{eq:foundation_KD}, with state-of-the-art GAN compression methods\cite{wang2020gan, liu2021content, li2021revisiting, kang2022information}. To distill the knowledge of $G^T$, \texttt{DiME} compares the outputs of $G^T$ and $G^S$ in the foundation embedding spaces (\ie, DINO, CLIP). As shown in \cref{table:quan_stylegan} and \cref{table:quan_sngan}, \texttt{DiME} outperforms the previous compression methods on various GAN architectures and datasets. Particularly, \texttt{DiME} improves FID scores by 1.63 compared to the state-of-the-art GAN compression methods in a setting where it reduces the FLOPS of StyleGAN2 for FFHQ by 11 times, with a compression rate of 90.73\%. Our experimental results show that \texttt{DiME} is highly effective for knowledge distillation.

Additionally, to investigate the effectiveness of distillation considering the characteristics of GANs (\ie, \texttt{NICKEL}) beyond direct knowledge distillation (\ie, \texttt{DiME}), we combine \texttt{NICKEL}, as described in \cref{NICKEL}, and \texttt{DiME}. \cref{table:quan_stylegan} and \cref{table:quan_sngan} show that \texttt{NICKEL \& DiME} further improves the performance over that of \texttt{DiME}. In \cref{table:quan_stylegan}, our method achieves the FID score of 15.93 by compressing StyleGAN2 93-fold. This compares to ITGC, which attains the FID score of 14.01 with a 24-fold compression. Furthermore, as shown in \cref{table:quan_stylegan}, our method not only obtains better computational efficiency but also shows superior performance compared to the state-of-the-art diffusion-model pruning method for the LSUN-Church dataset. These results indicate the continued significance of GAN compression research. As shown in \cref{table:quan_stylegan} and \cref{table:quan_sngan}, our method can be applied to various GAN architectures and show similar trends for various datasets.

\begin{table}[tb!]
\caption{Quantitative results of extremely compressed StyleGAN2. We compare the performance of various compression methods on StyleGAN2 for FFHQ at compression rate = 99.69\%. Previous methods often suffer from severe performance degradation due to the imbalance between $G^S$ and $D^S$ when GAN is extremely compressed. On the other hand, \texttt{NICKEL \& DiME} shows acceptable performance compared to other methods with high stability.}
\centering
\resizebox{\textwidth}{!}{
    \begin{tabular}{ccccccc}
    \toprule
    \multirow{1}{*}{Model}    & \multirow{1}{*}{Dataset}     & \multirow{1}{*}{Method} & \#params. & \multirow{1}{*}{FLOPs} & Compression rate & \multirow{1}{*}{FID$\downarrow$} \\  \midrule
    \multirow{5}{*}{StyleGAN2} & \multirow{5}{*}{\shortstack{FFHQ\\(256$\times$256)}}     & Full model \cite{karras2020training} & 24.77M               & 14.90B           & -     & 4.02        \\\cmidrule(l){3-7}
                              &                              & GS \cite{wang2020gan}           & \multirow{4}{*}{2.35M}     & \multirow{4}{*}{0.05B}         & \multirow{4}{*}{99.69\%}       & 184.33        \\
                              &                              & CAGC \cite{liu2021content}       &     &          &        & 186.61        \\
                              &                              & ITGC \cite{kang2022information}   &     &           &        & 164.92        \\
                              &                              & \texttt{NICKEL \& DiME}            &                       &        &           & \textbf{29.38}        \\ \bottomrule
    \end{tabular}
}
\label{table:highsparsity}
\end{table}
 
For in-depth investigations, we compare the FID, Precision, and Recall performance of compression methods at various compression rates. \cref{fig:performance} indicates the FID, Precision, and Recall scores at each compression rate for StyleGAN2 on FFHQ dataset. In \cref{fig:fid}, \texttt{NICKEL \& DiME} outperforms previous methods at all reported compression rates. Furthermore, at a compression rate of 74.96\%, \texttt{NICKEL \& DiME} shows only a 9.68\% performance degradation (FID: 4.42), compared to the full model (FID:4.03). Remarkably, our method shows significant gaps with previous methods as compression rates increase, thanks to the improved stability. \cref{fig:precision} shows precision scores, which indicate the fidelity of generated images. We observe that the precision scores of the compressed models are higher than those of the full model. While we observe a deterioration of precision scores with increasing compression rates, \texttt{NICKEL \& DiME} maintains the precision scores comparable to the full model, even at high compression rates. Moreover, \texttt{NICKEL \& DiME} shows precision scores comparable to the precision score of the full model up to a compression rate of 98.92\%. \cref{fig:recall} shows recall scores, indicating the diversity of the generated images. We observe that, unlike precision scores, the recall scores of the compressed models decrease significantly with increasing compression rates. Still, \texttt{NICKEL \& DiME} maintains better diversity compared to the other compression methods. We provide performance comparison with recent various metrics in the supplementary for more comprehensive analysis.

\begin{figure}[tb!]
  \centering
  \begin{subfigure}{0.45\linewidth}
    \includegraphics[height=5.5cm]{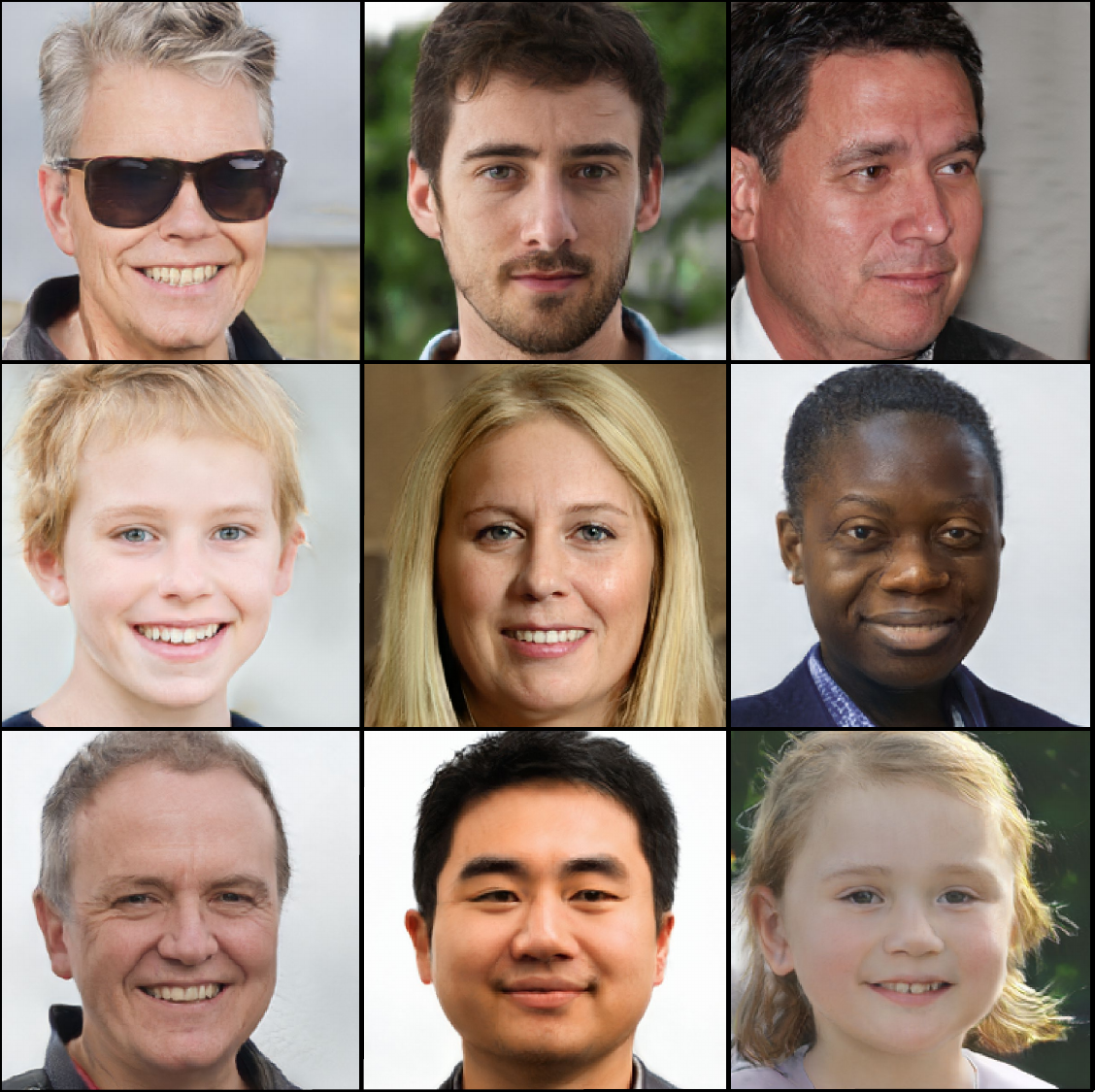}
    \caption{compression rate = 90.73\%}
    \label{fig:vis_ffhq70}
  \end{subfigure}
  \hfill
  \begin{subfigure}{0.45\linewidth}
    \includegraphics[height=5.5cm]{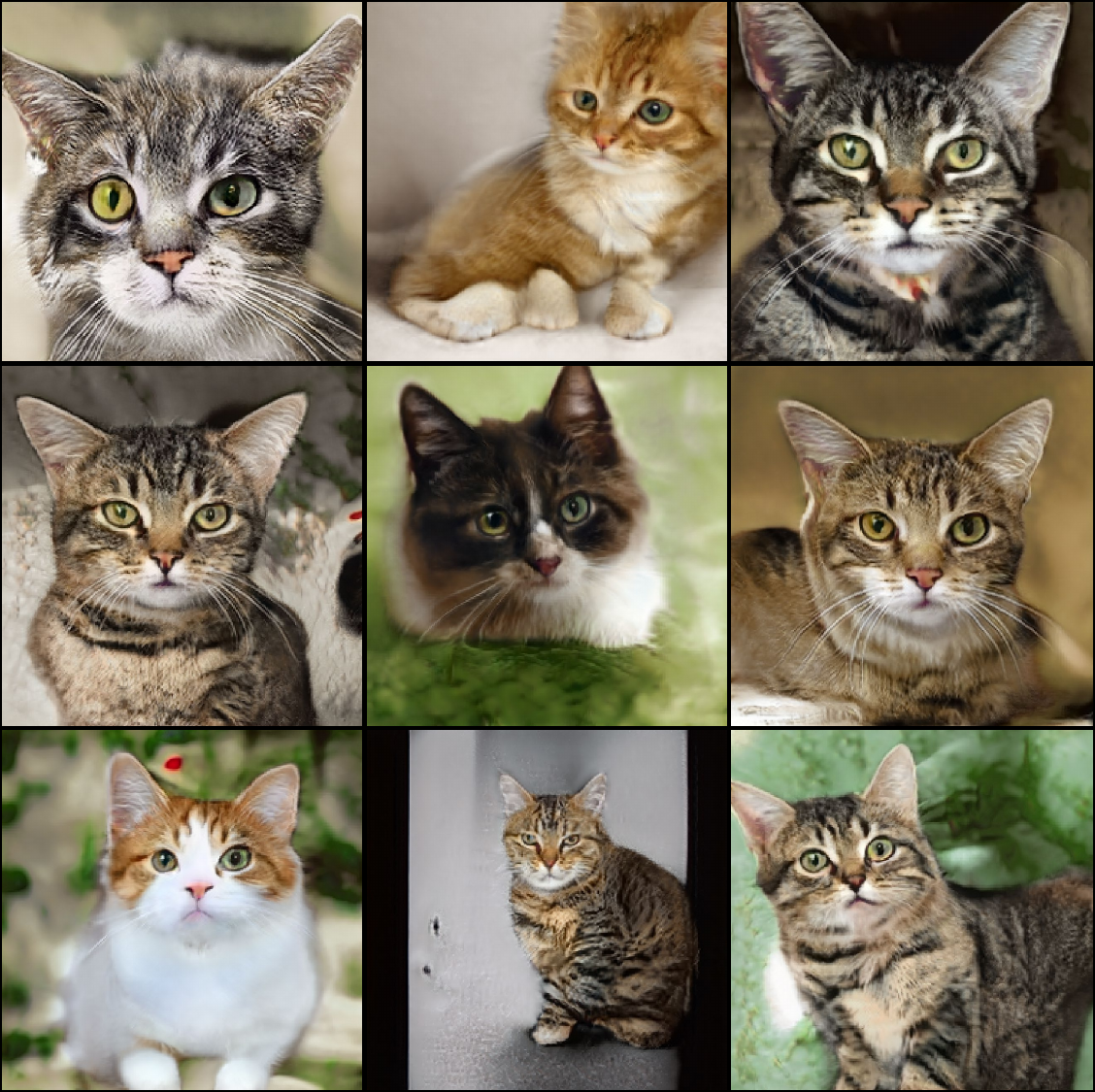}
    \caption{compression rate = 90.73\%}
    \label{fig:vis_cat70}
  \end{subfigure}  
  \caption{Visualization of images generated by compressed StyleGAN2 on FFHQ and LSUN-CAT. (a) shows the visual quality of StyleGAN2 compressed by \texttt{NICKEL \& DiME} on FFHQ at compression rate = 90.73\%. (b) shows the visual quality of StyleGAN2 compressed by \texttt{NICKEL \& DiME} on LSUN-CAT at compression rate = 90.73\%.}
  \label{fig:vis_70}
\end{figure}

As shown in \cref{fig:stability}, \texttt{NICKEL \& DiME} mitigates the imbalance between $G^S$ and $D^S$ by considering $D^S$ during knowledge distillation. \cref{fig:diseq} and \cref{fig:eq} respectively show the logits of $D^S$ for ITGC and \texttt{NICKEL \& DiME}. In contrast to the ideal training of GAN where the logits of the discriminator should be close to 0, ITGC shows significant performance degradation due to the imbalance between $G^S$ and $D^S$ during training. Particularly, as the compression on generator intensifies, the imbalance between $G^S$ and $D^S$ becomes more pronounced. On the other hand, \texttt{NICKEL \& DiME} alleviates this imbalance. Even at a compression rate of 98.92\%, our method maintains a better equilibrium compared to ITGC's at the compression rate of 90.73\%. \cref{fig:convergence} shows the convergence of FID scores, indicating stable convergence of our method compared to the other alternatives. It is noteworthy that our method shows stable convergence even under extreme compression rates. As shown in \cref{table:highsparsity}, at an extreme compression rate of 99.69\%, other methods fail to achieve stable learning due to the breakdown of Nash equilibrium between the highly compressed generator and discriminator. In contrast, our method not only shows stable convergence but also achieves reasonable performance, even with a 321-fold compression. \cref{section:visual} shows the visual quality of this scenario.

\begin{figure}[tb!]
  \centering
  \begin{subfigure}{0.45\linewidth}
    \includegraphics[height=5.5cm]{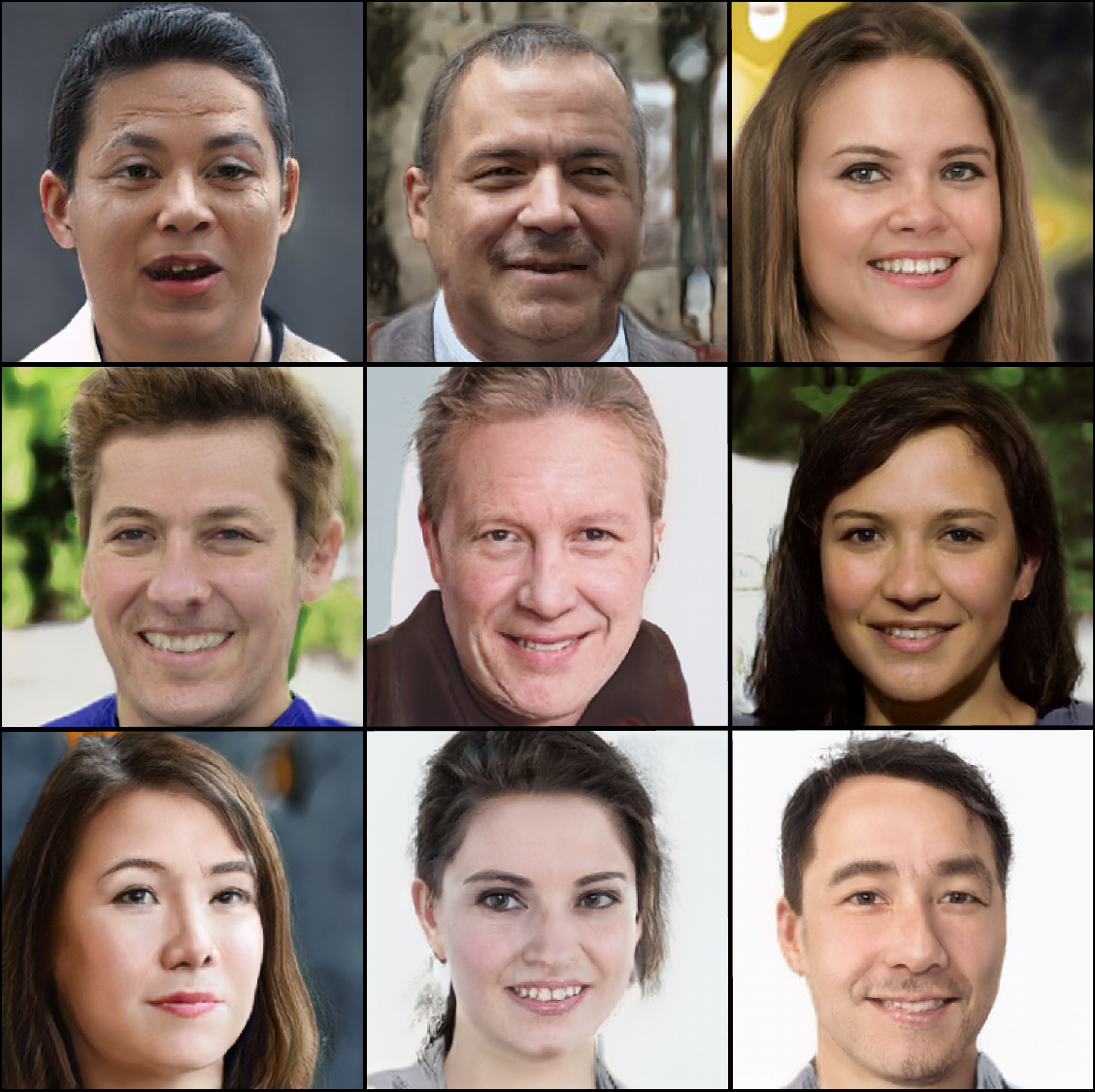}
    \caption{compression rate = 98.92\%}
    \label{fig:vis_ffhq90}
  \end{subfigure}
  \hfill
  \begin{subfigure}{0.45\linewidth}
    \includegraphics[height=5.5cm]{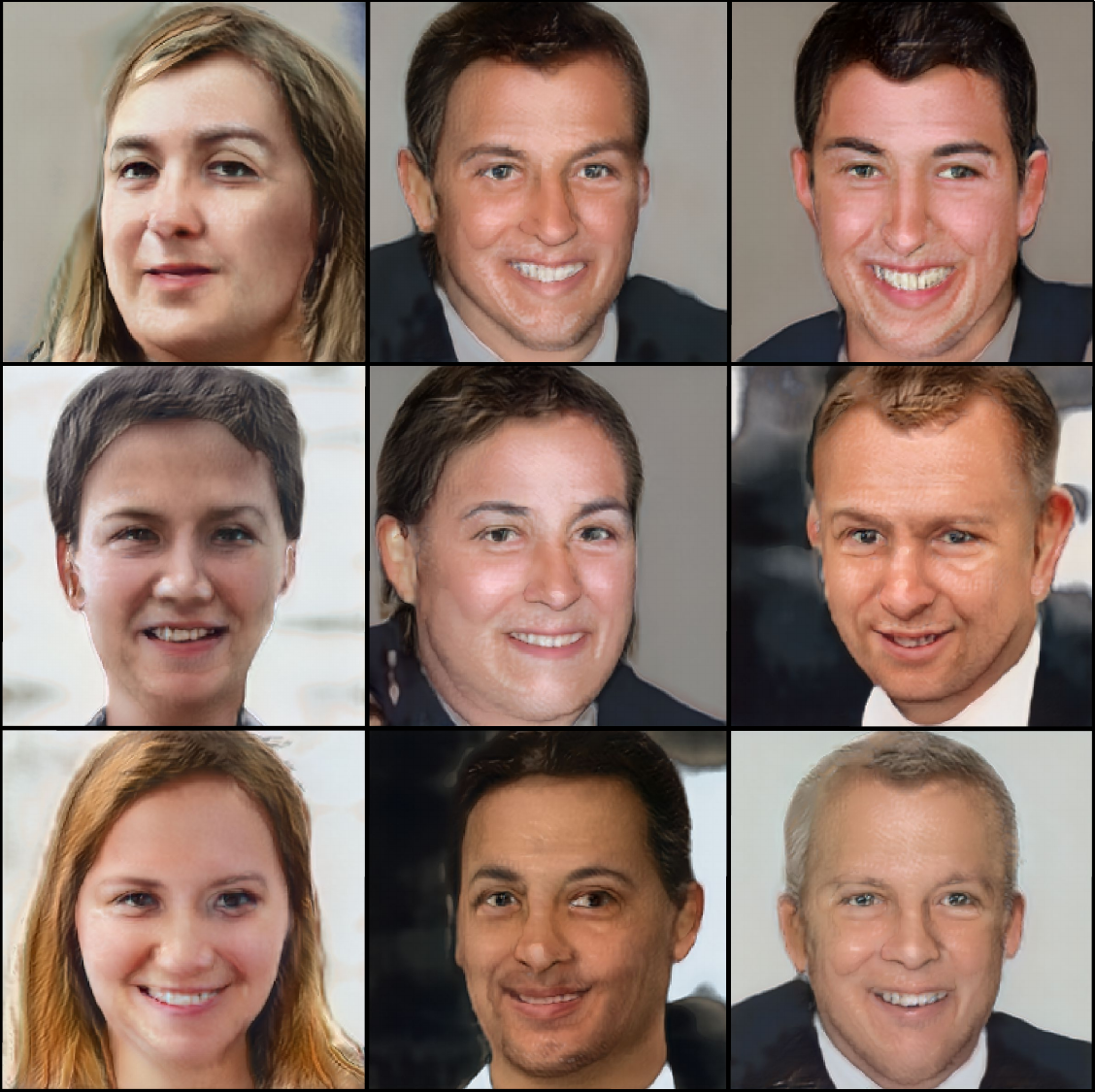}
    \caption{compression rate = 99.69\%}
    \label{fig:vis_ffhq95}
  \end{subfigure}  
  \caption{Visualization of extremely compressed StyleGAN2 on FFHQ. (a) indicates the visual quality of StyleGAN2 compressed by \texttt{NICKEL \& DiME} at compression rate = 98.92\%. (b) indicates the visual quality of StyleGAN2 compressed by \texttt{NICKEL \& DiME} at compression rate = 99.69\%. \texttt{NICKEL \& DiME} shows acceptable visual quality even at extreme compression rates.}
  \label{fig:vis_high}
\end{figure}

\subsection{Visualization of a Compression Factor of 11, 92, and 321.}
\label{section:visual}
In \cref{fig:vis_70}, we show generated images for FFHQ and LSUN-CAT datasets using StyleGAN2 at a compression rate of 90.73\%. Our method shows not only high visual quality but also the ability to generate diverse images. In \cref{fig:vis_high}, we visualize generated images at high compression rates. At a compression rate of 98.92\%, our method shows visual quality that is not significantly degraded. Moreover, even at a compression rate of 99.69\%, our method shows reasonable visual quality with diverse images.

\subsection{Ablation Study}
\label{sec:ablation}

\smallskip
\noindent\textbf{CLIP and DINO Embeddings.}
In \cref{table:ablation}, the CLIP embedding w/o global indicates using only CLIP as the embedding kernel for knowledge distillation. We observe significant challenges in achieving stable knowledge distillation when using only CLIP. In contrast, when using only DINO as the embedding kernel, DINO embedding w/o global, we observe stable convergence and achieve the FID of 20.75. In addition, we observe that although the CLIP embedding space may pose challenges in achieving stable knowledge distillation, combining it with the DINO embedding space could lead to slight performance improvements. 

\smallskip
\noindent\textbf{Utilization of Global Features.} 
The objective of KD is to match the population distributions between $G^T$ and $G^S$. However, due to the batch size, we can only match the sample distributions. Hence, a sampling error $\epsilon_{KD}$ may occur in the KD loss, which is bounded by the sum of the sampling errors of $G^T$ and $G^S$:
\begin{equation}
\epsilon_{KD} < \epsilon_{teacher} + \epsilon_{student}
\label{sampling_error}
\end{equation}
Fortunately, unlike $G^S$, the distribution of $G^T$ is fixed. Therefore, by precomputing the statistics--- referred to as global features ---through infinite sampling, we can achieve an infinitesimal sampling error $\epsilon_{teacher}$. As shown in \cref{table:ablation}, we find that utilizing global features leads to performance enhancement. 
In fact, this resembles the MMD critic, which is a stable metric for learning the distribution. Santos \etal \cite{santos2019learning} and Yeo \etal \cite{yeo2023can} noted that pretrained neural networks can be considered as characteristic kernels, and reducing the discrepancy of the mean between extracted features can be seen as the MMD critic. In this vein, \texttt{DiME} can be considered to stably match distributions between two generators.

\begin{table}[tb!]
\caption{Ablation study results in \texttt{NICKEL \& DiME}. }
\centering
\footnotesize
{
    \centering
    \begin{tabular}{ccccccc}
    \toprule
    \multirow{2}{*}{Name} &  \multicolumn{2}{c}{Model}    & \multirow{2}{*}{Global features} & \multirow{2}{*}{\texttt{NICKEL}}    & \multirow{2}{*}{FID$\downarrow$}     \\ \cline{2-3} 
                          &   DINO         & CLIP         &                                  &                         &                            \\ \midrule
    CLIP embedding w/o global           &                & \checkmark   &                                  &                         & 152.15                                \\     
    DINO embedding w/o global            &\checkmark      &              &                                  &                         & 20.75                          \\
    DINO embedding      &\checkmark      &              & \checkmark                       &                         & 19.60                          \\    
    \texttt{DiME}                  &\checkmark      & \checkmark   & \checkmark                       &                         & 18.32                          \\
    \texttt{NICKEL \& DiME}       &\checkmark      & \checkmark   & \checkmark                       & \checkmark              & \textbf{15.93}                           \\ \bottomrule
    \end{tabular}
}
\label{table:ablation}
\end{table}

\section{Limitations}
Our method shows excellent performance via distribution matching, yet it tends to focus on the fidelity of generated images. In fact, every method experiences significant degradation in recall performance, even at low compression rates (\cref{fig:performance}). Furthermore, there still remains the imbalance between the generator and discriminator at extreme compression rates, which incurs significant performance degradation. Thus, it is an interesting research direction to develop methods that are capable of maintaining diversity and stability when compressing generative models at extreme compression rates.

\section{Conclusion}
\label{section:conclusion}
In this paper, we propose Distribution Matching for Efficient compression (\texttt{DiME}) and Network Interactive Compression via Knowledge Exchange and Learning (\texttt{NICKEL}) that set a new standard of the performance in GAN compression. 
\texttt{DiME} matches the distributions between the teacher generator and student generator by using the maximum mean discrepancy (MMD) as a loss function. For better matching, we harness the power of the pretrained foundation model and use it as embedding kernels in MMD loss for knowledge distillation. \texttt{DiME} can compress StyleGAN2 with the FID of 4.02 by 20 times while maintaining reasonable performance with the FID of 11.25, achieving the state-of-the-art performance in all compression rates. \texttt{NICKEL} further enhances the performance by providing better feedback to the student generator from the discriminator. Combining these two, \texttt{NICKEL \& DiME} successfully compresses StyleGAN2 by 92 times while maintaining the FID score of 15.93. Thanks to its enhanced stability, \texttt{NICKEL \& DiME} allows us to compress StyleGAN2 by up to 99.69\% (321 times smaller) while maintaining reasonable performance, which is not possible for existing methods.

\clearpage
\section*{Acknowledgments}
This work was supported by the National Research Foundation of Korea (NRF) grant funded by the Korea government (MSIT) (No.2022R1C1C1008496), Institute of Information \& communications Technology Planning \& Evaluation (IITP) grant funded by the Korea government (MSIT) (No.RS-2020-II201336, Artificial Intelligence graduate school support (UNIST), No.RS-2021-II212068, Artificial Intelligence Innovation Hub, RS-2022-II220959, (Part 2) Few-Shot Learning of Causal Inference in Vision and Language for Decision Making, RS-2022-II220264, Comprehensive Video Understanding and Generation with Knowledge-based Deep Logic Neural Network). We also thank the supercomputing resources of the UNIST Supercomputing Center.

\bibliographystyle{splncs04}
\bibliography{egbib}
\end{document}


\title{Nickel and Diming Your GAN: A Dual-Method Approach to Enhancing GAN Efficiency via Knowledge Distillation \\-Supplementary Materials-}

\titlerunning{Nickel and Diming Your GAN}

\author{Sangyeop Yeo\orcidlink{0000-0002-5305-3443} \and
Yoojin Jang\orcidlink{0000-0001-8150-3715} \and
Jaejun Yoo\textsuperscript{*}\orcidlink{0000-0001-5252-9668}}

\authorrunning{S. Yeo et al.}

\institute{LAIT, Ulsan National Institute of Science and Technology (UNIST) \email{sangyeop377@gmail.com, \{softjin, jaejun.yoo\}@unist.ac.kr \\(*: corresponding author)}}

\maketitle

In this document, we provide additional analysis, and visualization for the Anonymous ECCV 2024 submission, "Nickel and Diming Your GAN: A Dual-Method Approach to Enhancing GAN Efficiency via Knowledge Distillation". We compare GAN-Slimming (GS)\cite{wang2020gan}, Content-Aware GAN Compression (CAGC)\cite{liu2021content}, Information-Theoretic GAN Compression (ITGC) \cite{kang2022information}, and \texttt{NICKEL\&DiME} (Ours). Our code is available at:\\\href{https://lait-cvlab.github.io/Nickel_and_Diming_Your_GAN/}{https://lait-cvlab.github.io/Nickel$\_$and$\_$Diming$\_$Your$\_$GAN/}.

\section{Quantitative Results}
To deeply investigate GAN compression methods, we evaluate them by using recent evaluation metrics: Density \& Coverage\cite{naeem2020reliable}, and Topological Precision \& Recall\cite{kim2024topp}. 
\begin{figure}[h!]
  \centering
  \begin{subfigure}{0.45\linewidth}
    \includegraphics[height=4.cm]{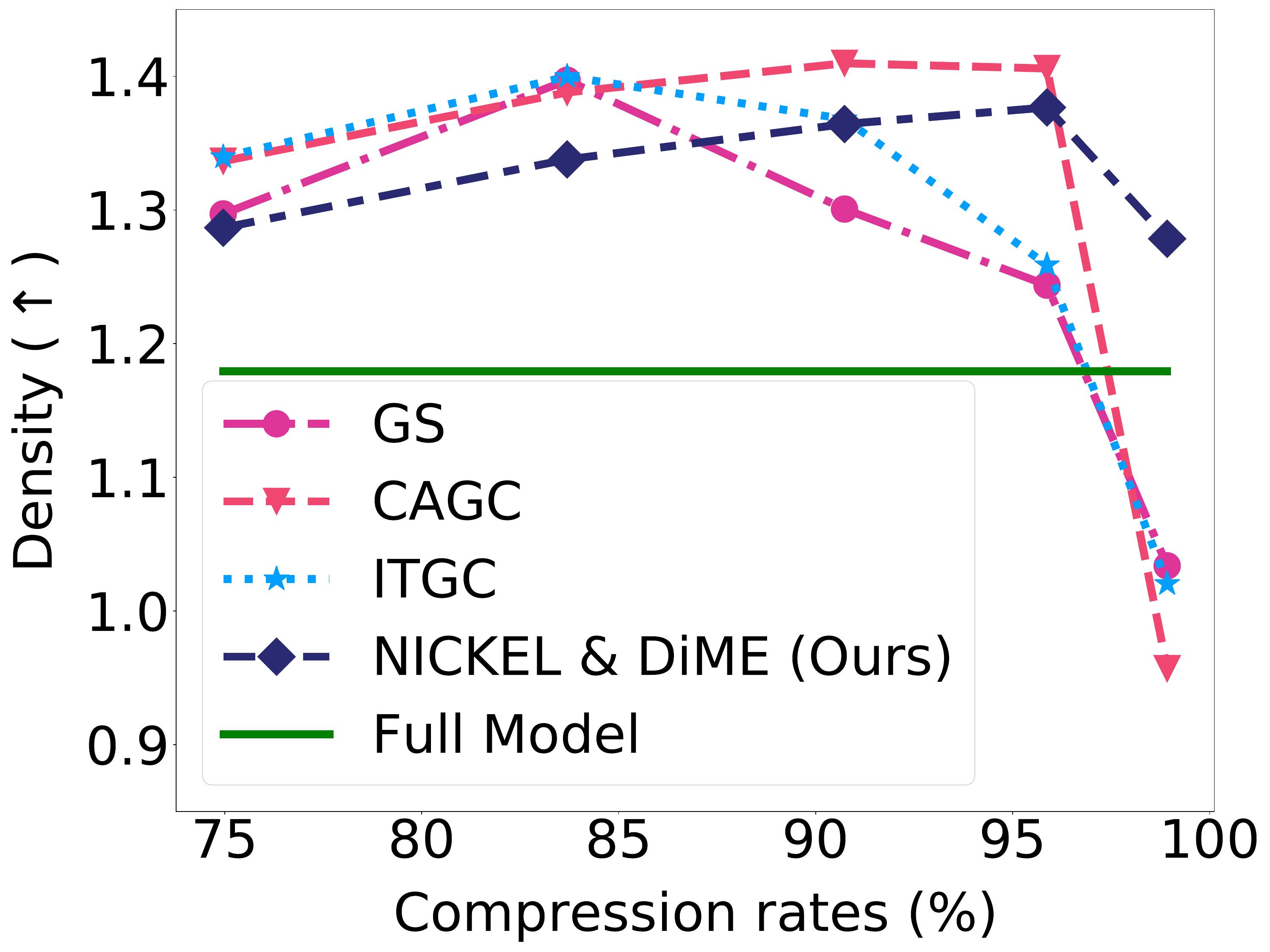}
    \caption{Comparison of Density}
    \label{fig:density}
  \end{subfigure}
  \hfill
  \begin{subfigure}{0.45\linewidth}
    \includegraphics[height=4.cm]{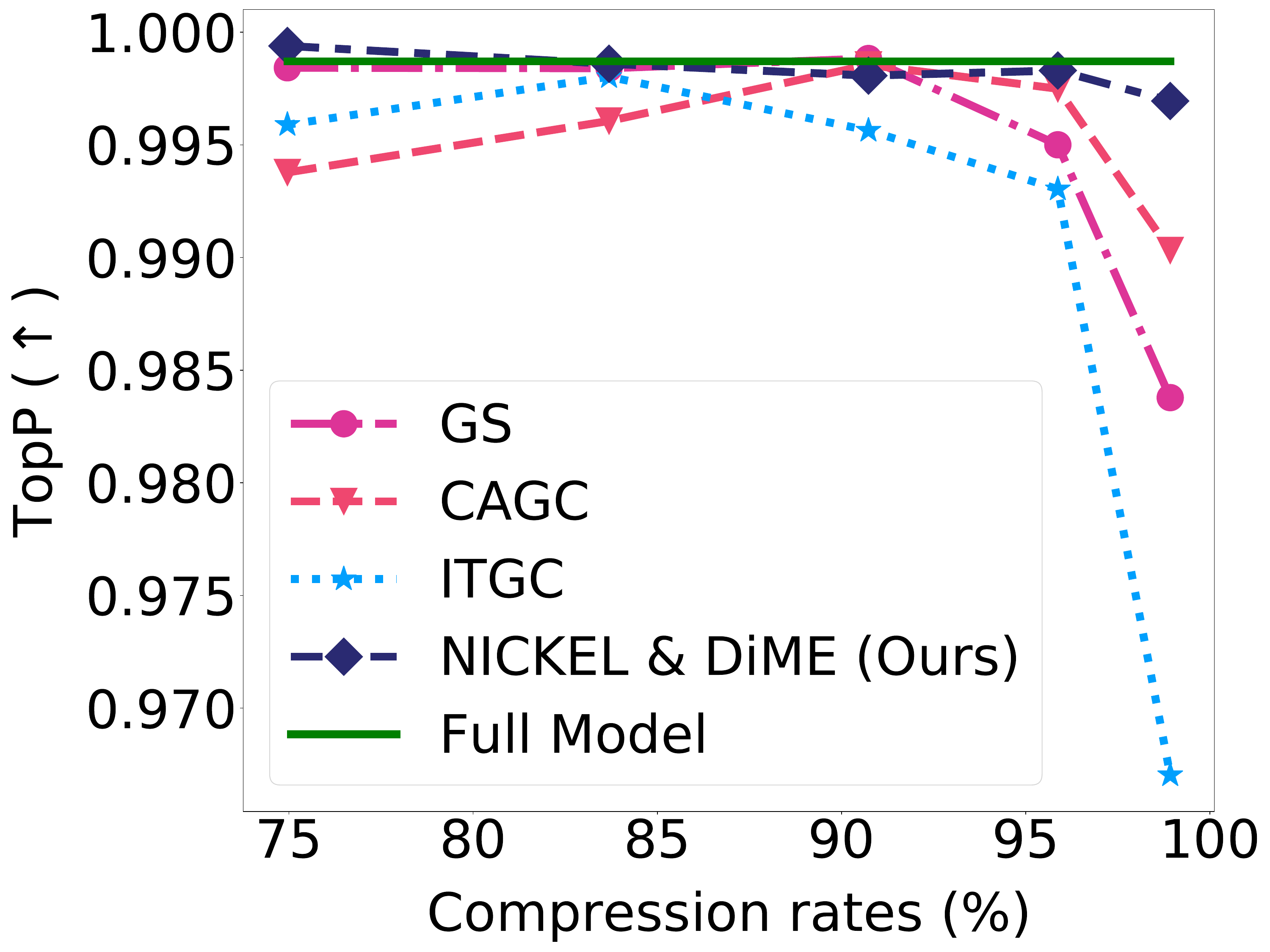}
    \caption{Comparison of Topological Precision}
    \label{fig:topp}
  \end{subfigure}  
  \caption{Comparison of Density and Topological Precision. (a) shows the Density scores that robustly evaluate the fidelity of the images to outliers. (b) shows the Topological Precision scores that measure the fidelity of the images topologically. (a) and (b) show a similar trend to Fig. 4b in the main manuscript.}
  \label{fig:fidelity}
\end{figure}

\begin{figure}[h!]
  \centering
  \begin{subfigure}{0.45\linewidth}
    \includegraphics[height=4.cm]{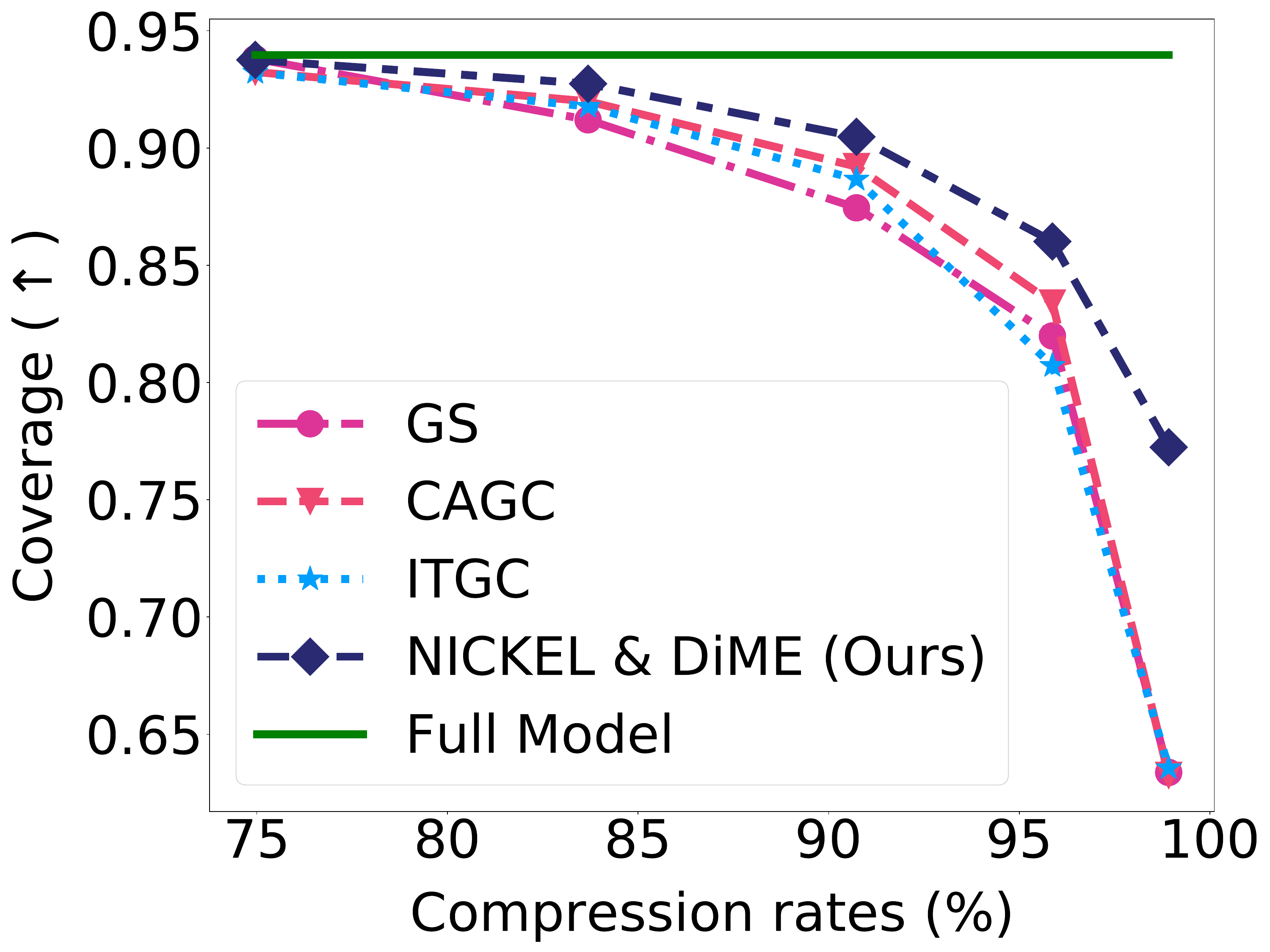}
    \caption{Comparison of Coverage}
    \label{fig:coverage}
  \end{subfigure}
  \hfill
  \begin{subfigure}{0.45\linewidth}
    \includegraphics[height=4.cm]{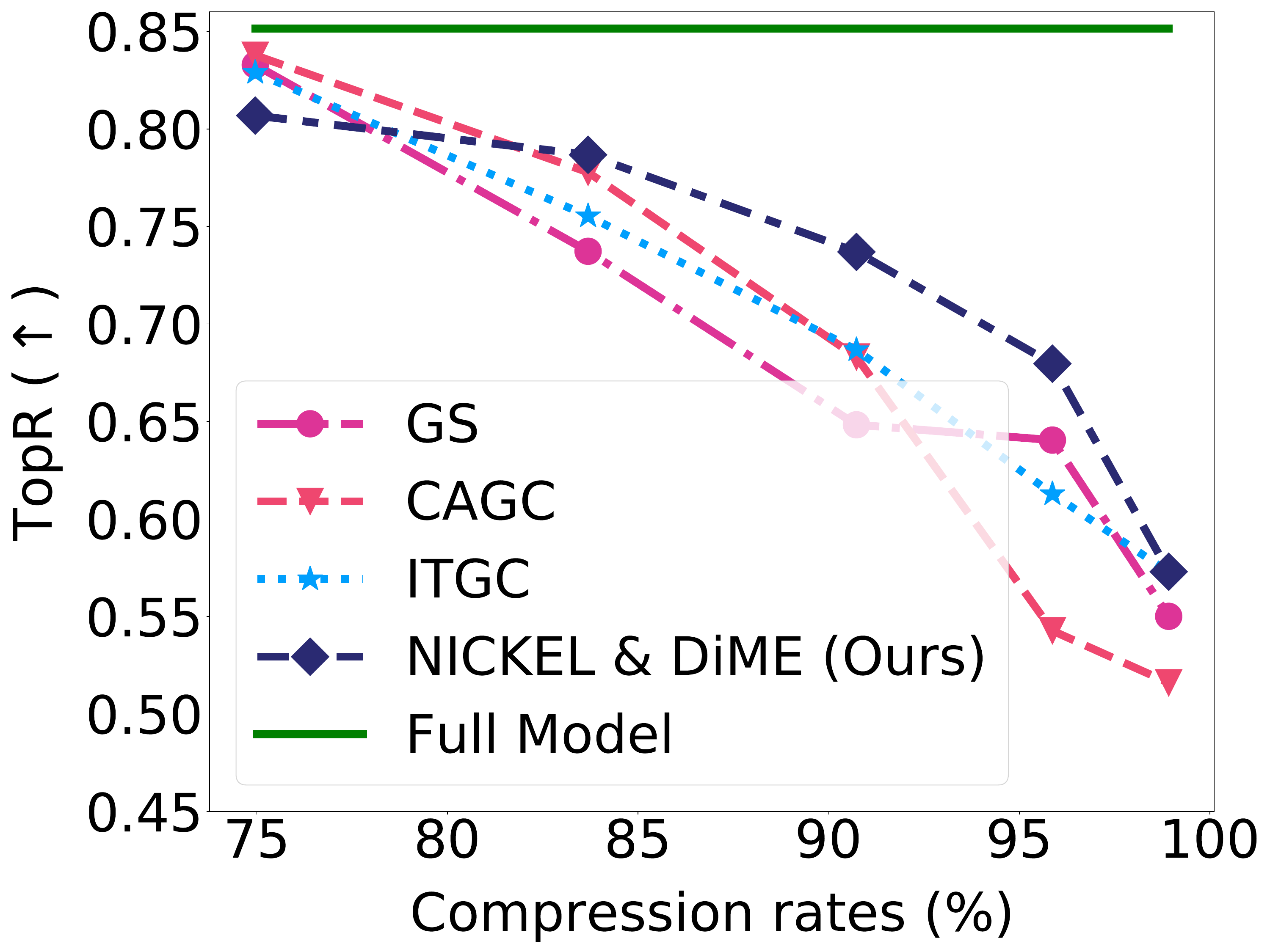}
    \caption{Comparsion of Topological Recall}
    \label{fig:topr}
  \end{subfigure}  
  \caption{Comparsion of Coverage and Topological Recall. (a) shows the Coverage scores that robustly evaluate the diversity of the images to outliers. (b) shows the Topological Recall scores that measure the diversity of the images topologically. (a) and (b) show a similar trend to Fig. 4c in the main manuscript.}
  \label{fig:diversity}
\end{figure}
\subsection{Evaluation metrics}
Density and Coverage \cite{naeem2020reliable} measure the fidelity and diversity of images respectively, robustly evaluating models against outliers when approximating the manifold of data using k-nearest neighbor. Topological Precision and Recall\cite{kim2024topp} provide evaluation while capturing significant features topologically and statistically, even in various scenarios such as outliers or mode drop situations.

\cref{fig:density} and \cref{fig:coverage} represent Density and Coverage, respectively, which robustly measure the fidelity and diversity of images generated against outliers. \cref{fig:topp} and \cref{fig:topr} represent Topological Precision and Topological Recall, respectively, which evaluate the fidelity and diversity of generated images topologically. In \cref{fig:density} and \cref{fig:topp}, which measure fidelity, we observe similar trends to Fig. 4b in the main manuscript. Additionally, In \cref{fig:coverage} and \cref{fig:topr}, which measure diversity, we observe similar trends to Fig. 4c in the main manuscript. In other words, \texttt{NICKEL\&DiME} consistently shows higher fidelity and diversity compared to other GAN compression methods.

\section{Qualitative Results}
\begin{figure}[h!]
  \centering
  \begin{subfigure}{\linewidth}
    \includegraphics[height=1.53cm]{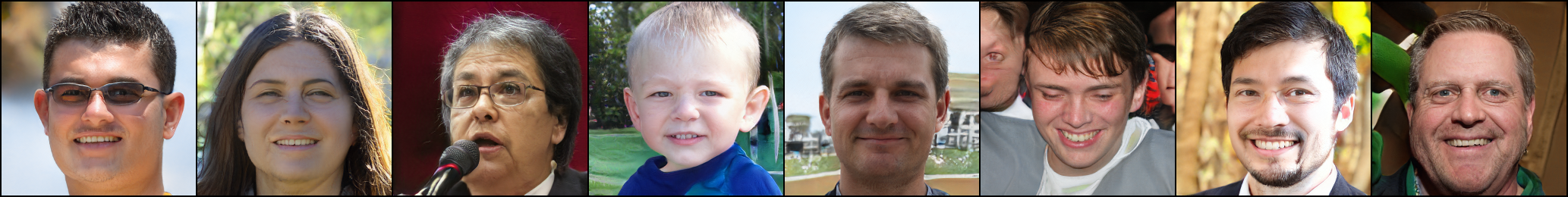}
    \caption{Full Model}
  \end{subfigure}
      \hspace{0.05\textwidth}

  \begin{subfigure}{\linewidth}
    \includegraphics[height=1.53cm]{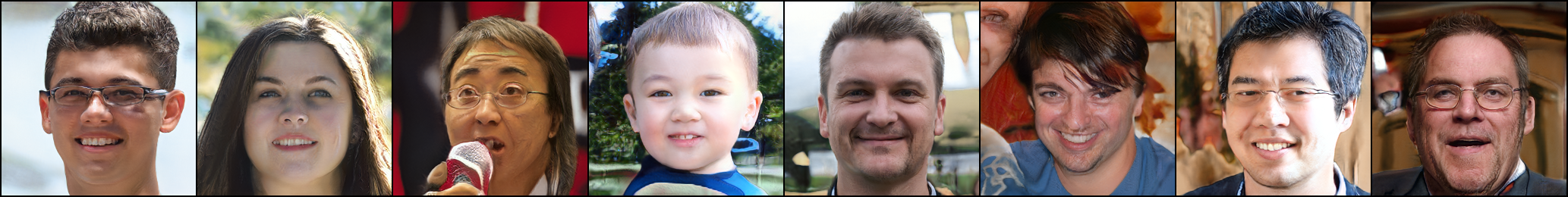}
    \caption{GAN-Slimming}
  \end{subfigure}
  \hspace{0.05\textwidth}
  
  \begin{subfigure}{\linewidth}
    \includegraphics[height=1.53cm]{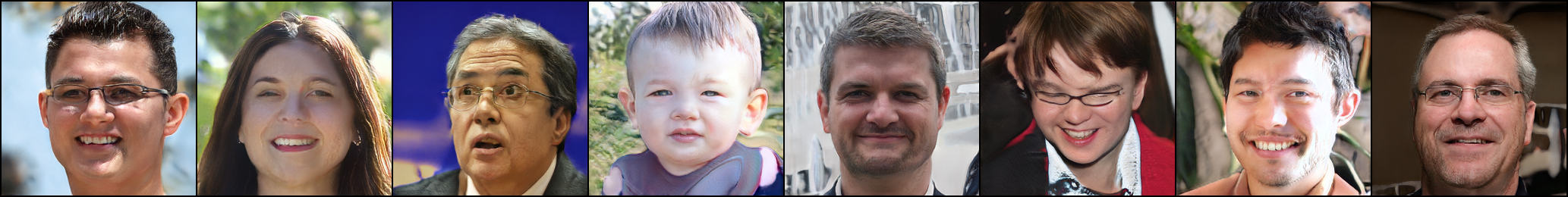}
    \caption{Content-Aware GAN Compression}
  \end{subfigure}
\hspace{0.05\textwidth}

  \begin{subfigure}{\linewidth}
    \includegraphics[height=1.53cm]{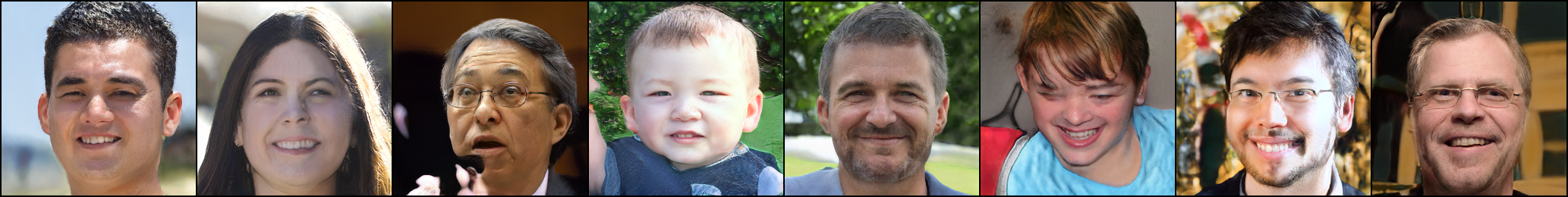}
    \caption{Information-Theoretic GAN Compression}
  \end{subfigure}
  \hspace{0.05\textwidth}
  
  \begin{subfigure}{\linewidth}
    \includegraphics[height=1.53cm]{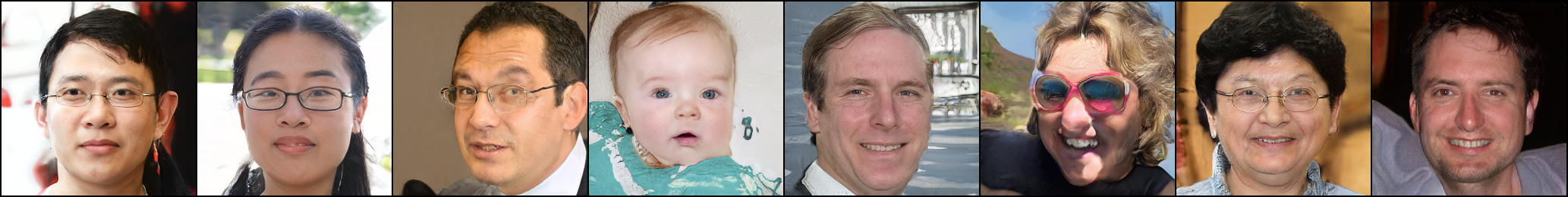}
    \caption{\texttt{DiME}}
  \end{subfigure}
  \hspace{0.05\textwidth}
  
  \begin{subfigure}{\linewidth}
    \includegraphics[height=1.53cm]{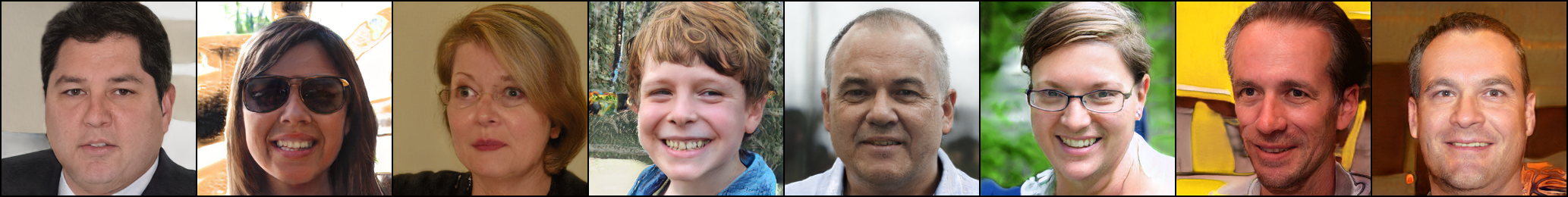}
    \caption{\texttt{NICKEL}}
  \end{subfigure}
  \hspace{0.05\textwidth}
  
  \begin{subfigure}{\linewidth}
    \includegraphics[height=1.53cm]{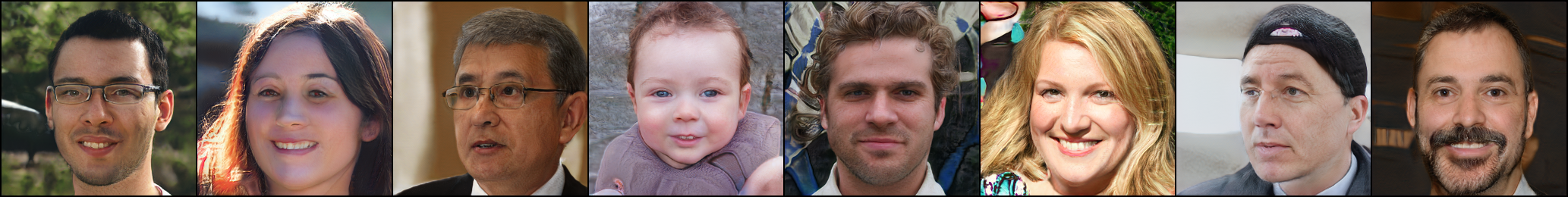}
    \caption{\texttt{NICKEL\&DiME}}
  \end{subfigure}  
  \caption{Visualization of generated images at a compression rate of 90.73\%. We show generated images from the full StyleGAN2 and compressed StyleGAN2 on the FFHQ dataset. (a), (b), (c), (d), (e), (f), and (g) represent the full model, GS, CAGC, ITGC, \texttt{DiME}, \texttt{NICKEL}, and \texttt{NICKEL\&DiME}, respectively.}
  \label{70}
\end{figure}

\begin{figure}[h!]
  \centering
  \begin{subfigure}{\linewidth}
    \includegraphics[height=1.53cm]{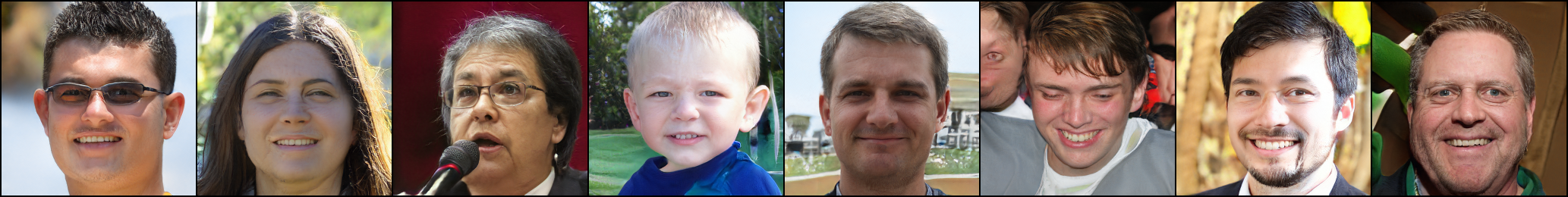}
    \caption{Full Model}
  \end{subfigure}
      \hspace{0.05\textwidth}

  \begin{subfigure}{\linewidth}
    \includegraphics[height=1.53cm]{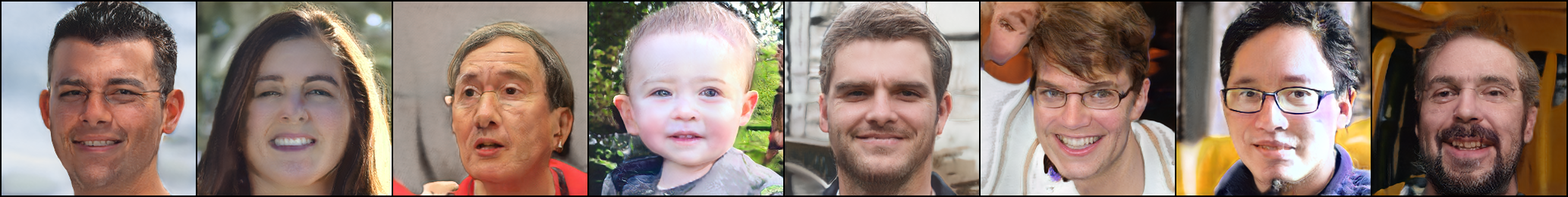}
    \caption{GAN-Slimming}
  \end{subfigure}
  \hspace{0.05\textwidth}
  
  \begin{subfigure}{\linewidth}
    \includegraphics[height=1.53cm]{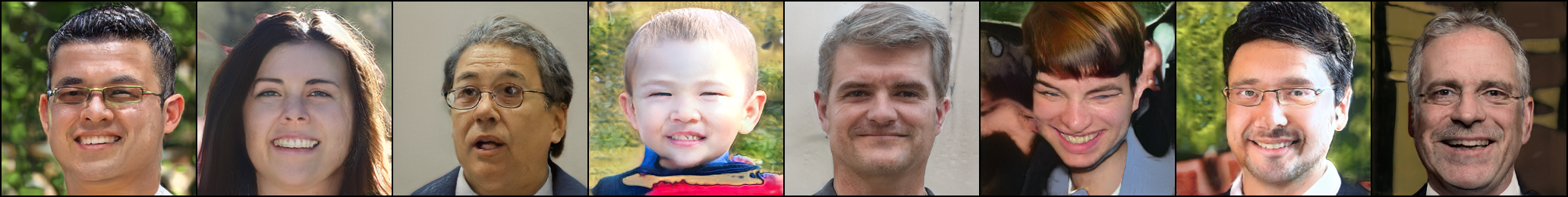}
    \caption{Content-Aware GAN Compression}
  \end{subfigure}
\hspace{0.05\textwidth}

  \begin{subfigure}{\linewidth}
    \includegraphics[height=1.53cm]{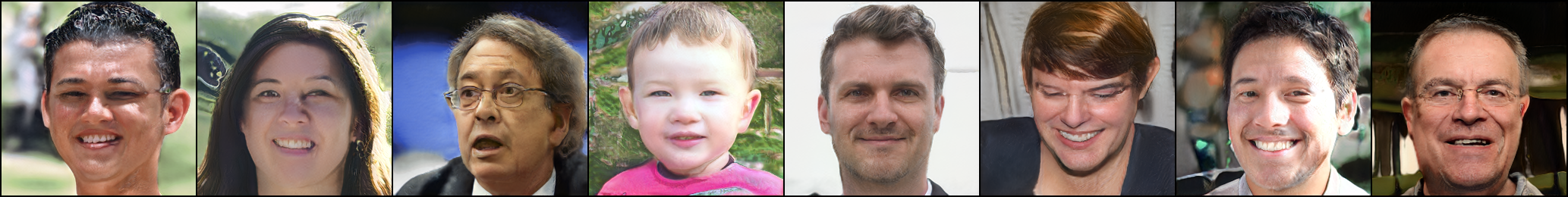}
    \caption{Information-Theoretic GAN Compression}
  \end{subfigure}
  \hspace{0.05\textwidth}
  
  \begin{subfigure}{\linewidth}
    \includegraphics[height=1.53cm]{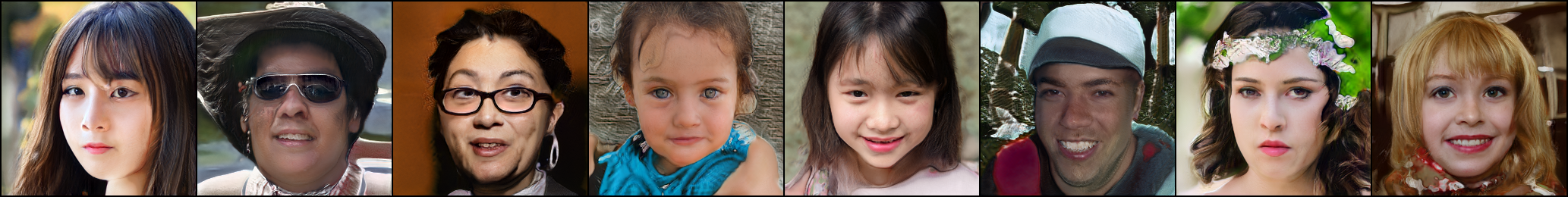}
    \caption{\texttt{DiME}}
  \end{subfigure}
  \hspace{0.05\textwidth}
  
  \begin{subfigure}{\linewidth}
    \includegraphics[height=1.53cm]{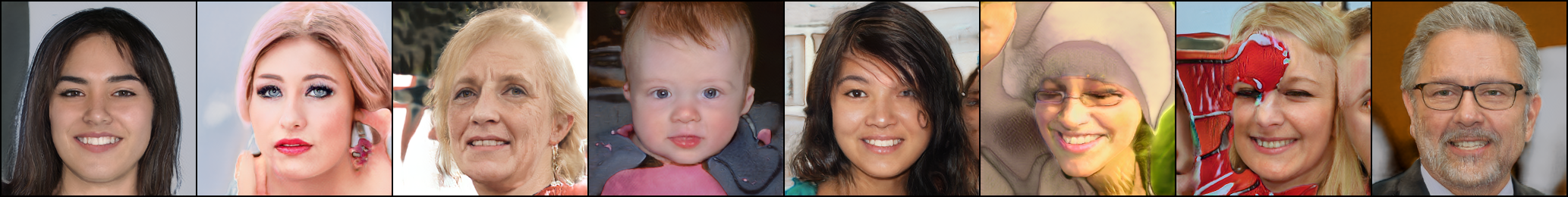}
    \caption{\texttt{NICKEL}}
  \end{subfigure}
  \hspace{0.05\textwidth}
  
  \begin{subfigure}{\linewidth}
    \includegraphics[height=1.53cm]{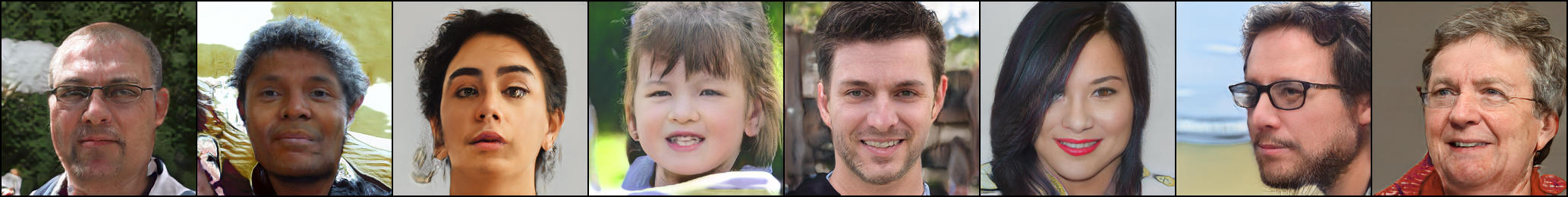}
    \caption{\texttt{NICKEL\&DiME}}
  \end{subfigure}  
  \caption{Visualization of generated images at a compression rate of 95.87\%. We show generated images from the full StyleGAN2 and compressed StyleGAN2 on the FFHQ dataset. (a), (b), (c), (d), (e), (f), and (g) represent the full model, GS, CAGC, ITGC, \texttt{DiME}, \texttt{NICKEL}, and \texttt{NICKEL\&DiME}, respectively.}
  \label{80}
\end{figure}

We provide the qualitative results of GAN compression methods at various compression rates. In \cref{70} and \cref{80}, we observe that previous GAN compression methods generate outputs similar to the full model for the same inputs. However, as demonstrated in Fig. 1c in the main manuscript, in unconditional GAN compression, the full model and the compressed model do not necessarily have to generate the same outputs. \cref{95} shows that the images generated by previous GAN compression methods have significantly deteriorated at extreme compression rates. On the other hand, our method shows reasonable visual quality.

\begin{figure}[h!]
  \centering
  \begin{subfigure}{\linewidth}
    \includegraphics[height=1.53cm]{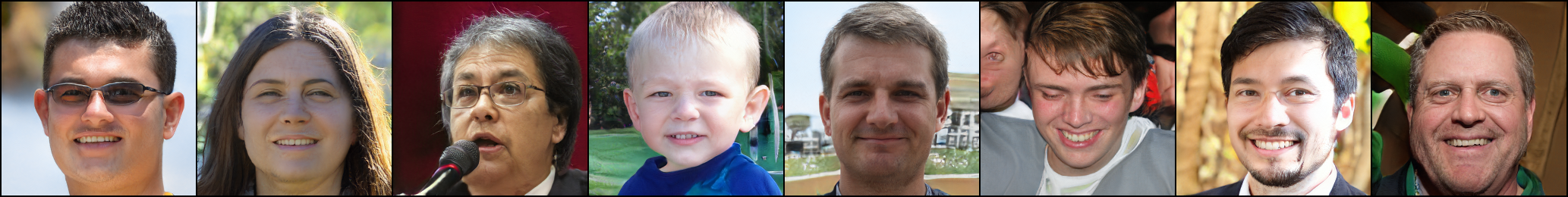}
    \caption{Full Model}
  \end{subfigure}
      \hspace{0.05\textwidth}

  \begin{subfigure}{\linewidth}
    \includegraphics[height=1.53cm]{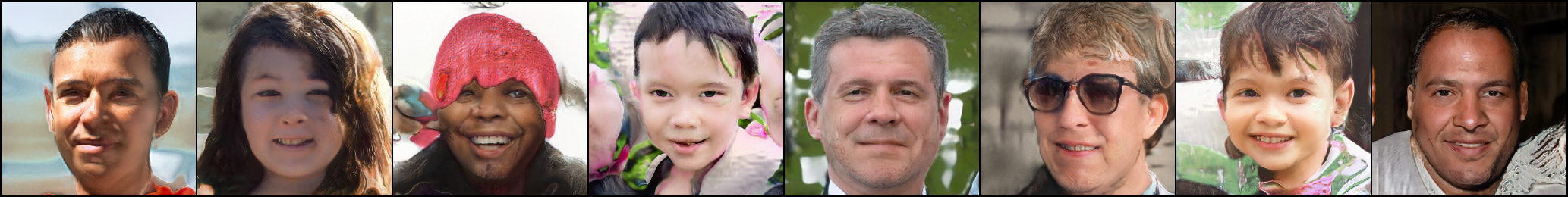}
    \caption{GAN-Slimming}
  \end{subfigure}
  \hspace{0.05\textwidth}
  
  \begin{subfigure}{\linewidth}
    \includegraphics[height=1.53cm]{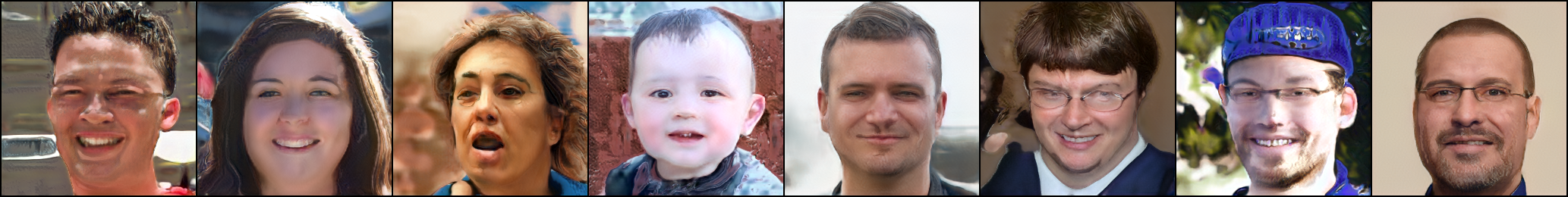}
    \caption{Content-Aware GAN Compression}
  \end{subfigure}
\hspace{0.05\textwidth}

  \begin{subfigure}{\linewidth}
    \includegraphics[height=1.53cm]{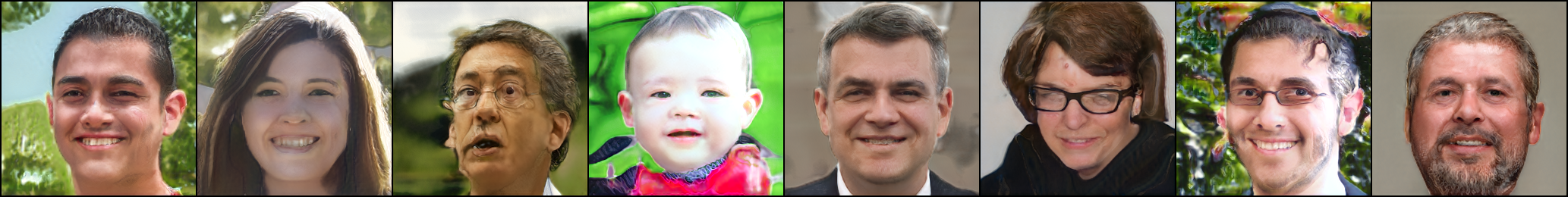}
    \caption{Information-Theoretic GAN Compression}
  \end{subfigure}
  \hspace{0.05\textwidth}
  
  \begin{subfigure}{\linewidth}
    \includegraphics[height=1.53cm]{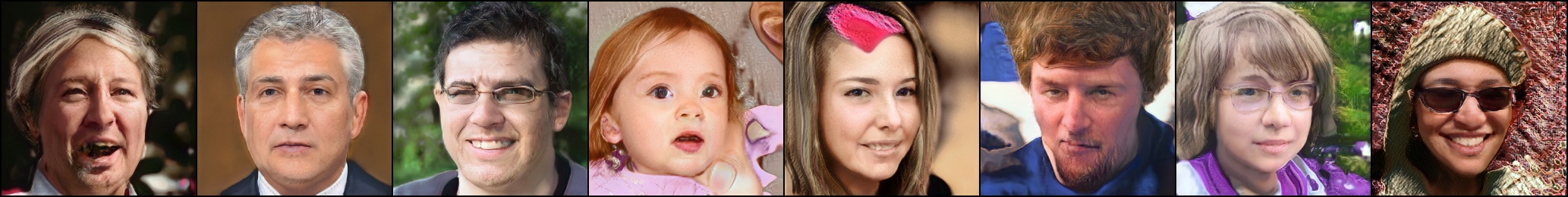}
    \caption{\texttt{DiME}}
  \end{subfigure}
  \hspace{0.05\textwidth}
  
  \begin{subfigure}{\linewidth}
    \includegraphics[height=1.53cm]{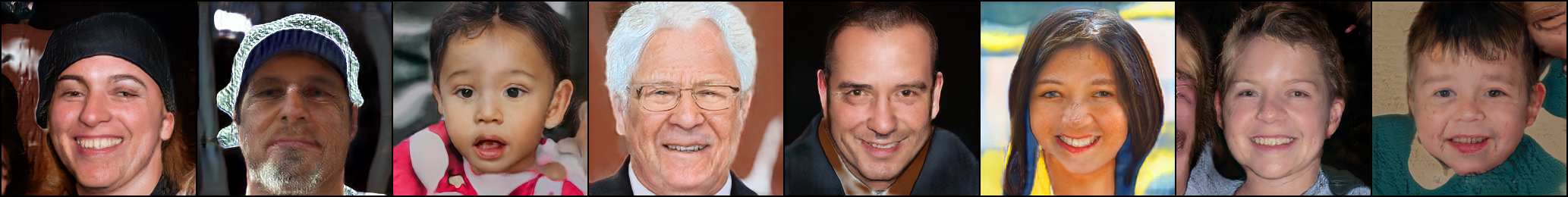}
    \caption{\texttt{NICKEL}}
  \end{subfigure}
  \hspace{0.05\textwidth}
  
  \begin{subfigure}{\linewidth}
    \includegraphics[height=1.53cm]{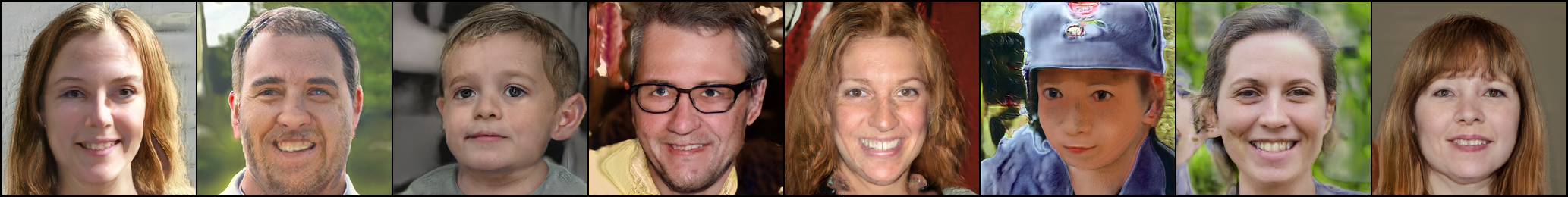}
    \caption{\texttt{NICKEL\&DiME}}
  \end{subfigure}  
  \caption{Visualization of generated images at a compression rate of 98.92\%. We show generated images from the full StyleGAN2 and compressed StyleGAN2 on the FFHQ dataset. (a), (b), (c), (d), (e), (f), and (g) represent the full model, GS, CAGC, ITGC, \texttt{DiME}, \texttt{NICKEL}, and \texttt{NICKEL\&DiME}, respectively.}
  \label{90}
\end{figure}

\begin{figure}[h!]
  \centering
  \begin{subfigure}{\linewidth}
    \includegraphics[height=1.53cm]{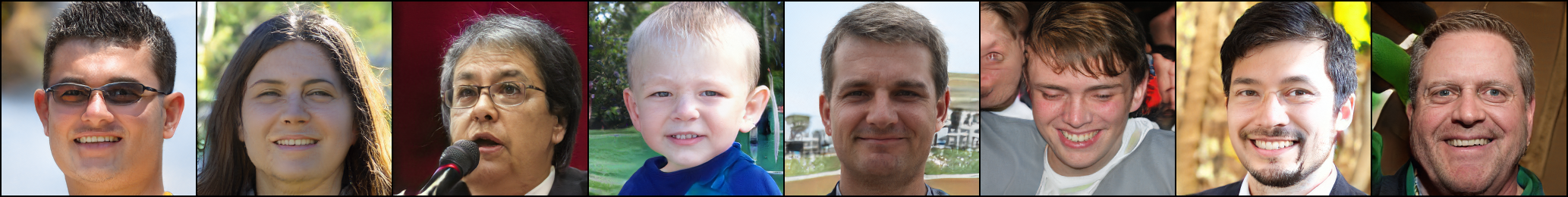}
    \caption{Full Model}
  \end{subfigure}
      \hspace{0.05\textwidth}

  \begin{subfigure}{\linewidth}
    \includegraphics[height=1.53cm]{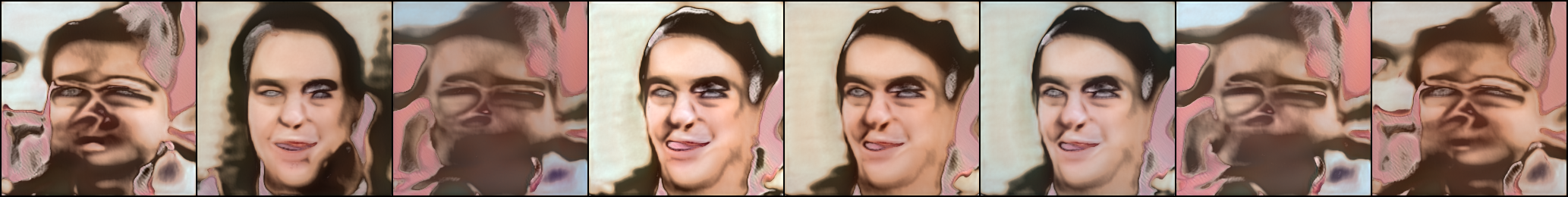}
    \caption{GAN-Slimming}
  \end{subfigure}
  \hspace{0.05\textwidth}
  
  \begin{subfigure}{\linewidth}
    \includegraphics[height=1.53cm]{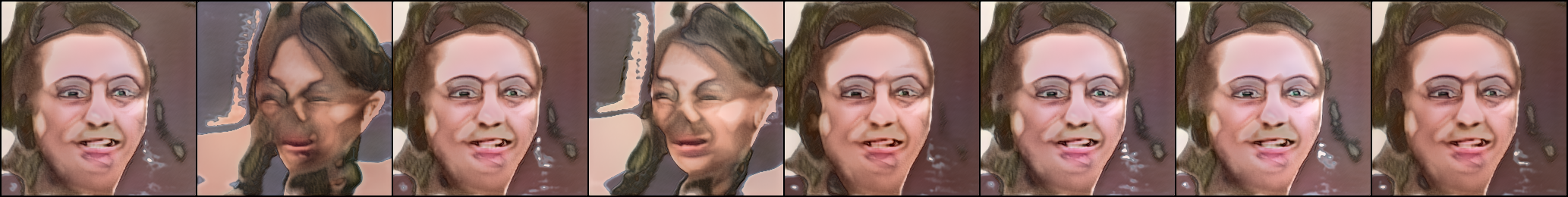}
    \caption{Content-Aware GAN Compression}
  \end{subfigure}
\hspace{0.05\textwidth}

  \begin{subfigure}{\linewidth}
    \includegraphics[height=1.53cm]{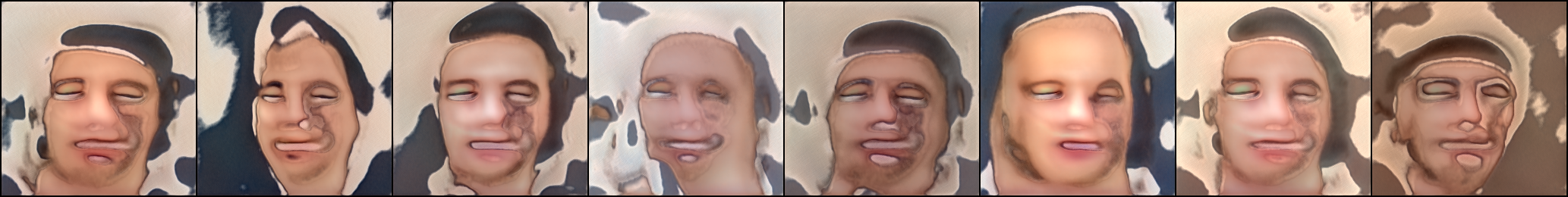}
    \caption{Information-Theoretic GAN Compression}
  \end{subfigure}
  \hspace{0.05\textwidth}
  
  \begin{subfigure}{\linewidth}
    \includegraphics[height=1.53cm]{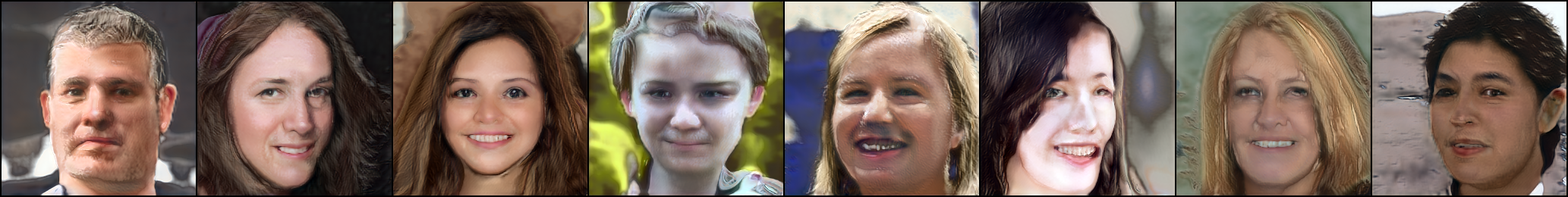}
    \caption{\texttt{NICKEL\&DiME}}
  \end{subfigure}  
  \caption{Visualization of generated images at a compression rate of 99.69\%. We show generated images from the full StyleGAN2 and compressed StyleGAN2 on the FFHQ dataset. (a), (b), (c), (d), and (e) represent the full model, GS, CAGC, ITGC, and \texttt{NICKEL\&DiME}, respectively.}
  \label{95}
\end{figure}

\begin{figure}[h!]
  \centering
  \begin{subfigure}{\linewidth}
    \includegraphics[height=1.53cm]{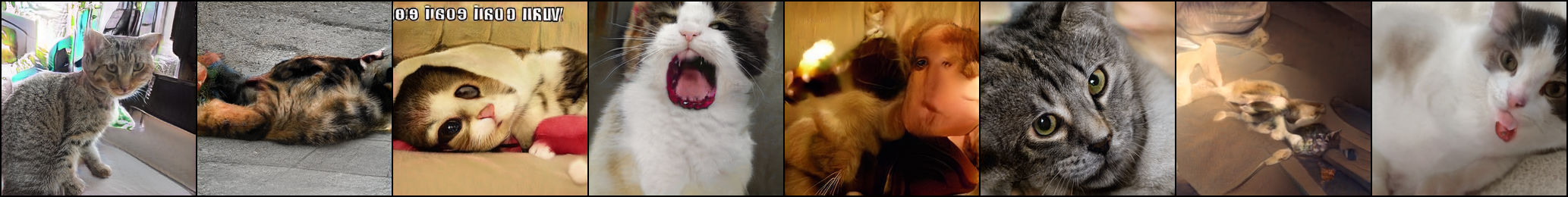}
    \caption{Full Model}
  \end{subfigure}
      \hspace{0.05\textwidth}

  \begin{subfigure}{\linewidth}
    \includegraphics[height=1.53cm]{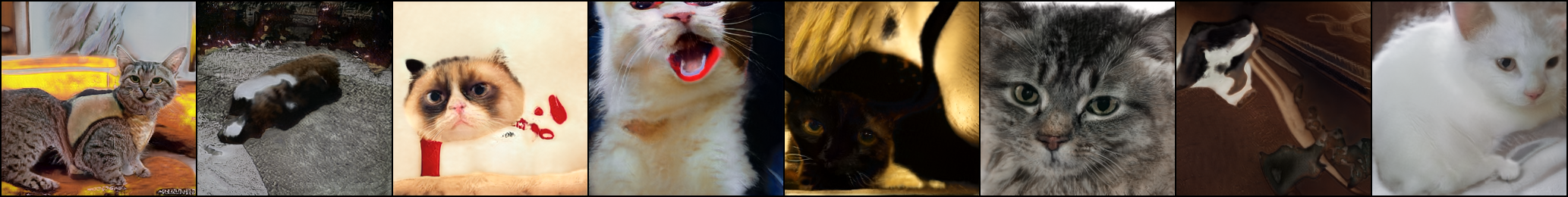}
    \caption{GAN-Slimming}
  \end{subfigure}
  \hspace{0.05\textwidth}
  
  \begin{subfigure}{\linewidth}
    \includegraphics[height=1.53cm]{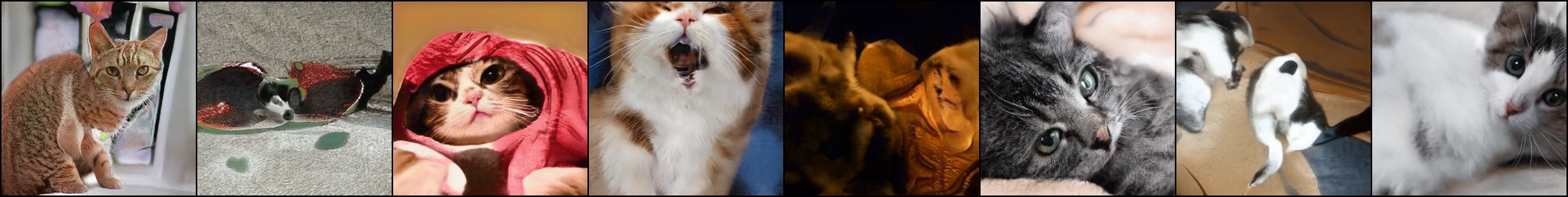}
    \caption{Content-Aware GAN Compression}
  \end{subfigure}
\hspace{0.05\textwidth}

  \begin{subfigure}{\linewidth}
    \includegraphics[height=1.53cm]{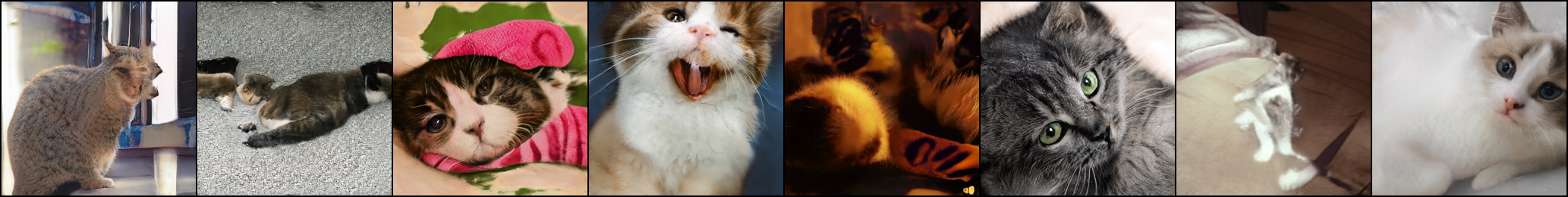}
    \caption{Information-Theoretic GAN Compression}
  \end{subfigure}
  \hspace{0.05\textwidth}

  \begin{subfigure}{\linewidth}
    \includegraphics[height=1.53cm]{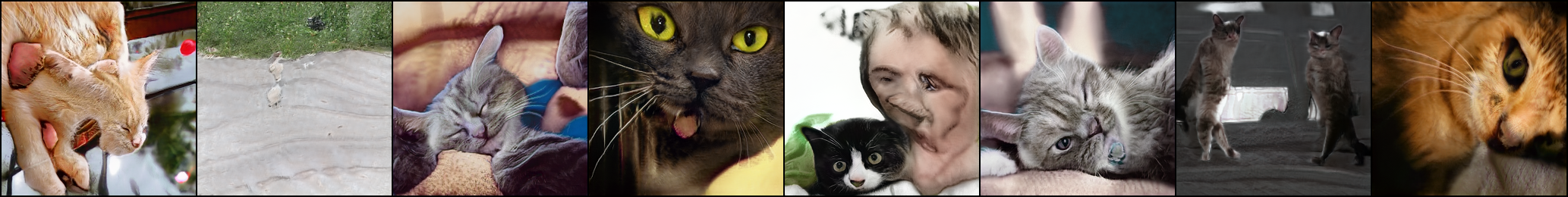}
    \caption{\texttt{DiME}}
  \end{subfigure}
  \hspace{0.05\textwidth}
  
  \begin{subfigure}{\linewidth}
    \includegraphics[height=1.53cm]{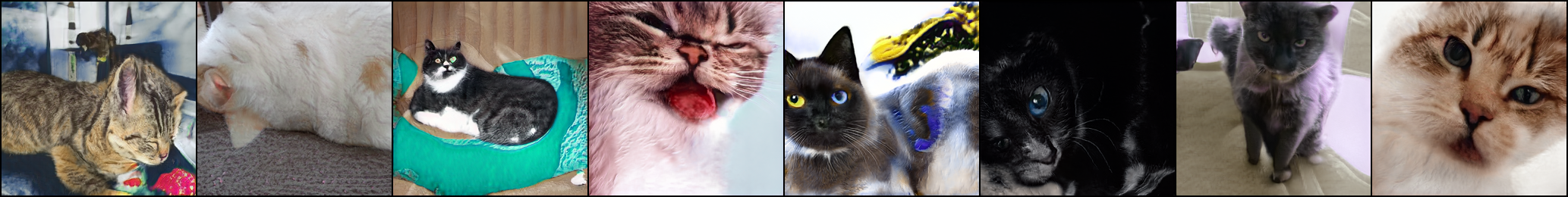}
    \caption{\texttt{NICKEL\&DiME}}
  \end{subfigure}  
  \caption{Visualization of generated images at a compression rate of 90.73\%. We show generated images from the full StyleGAN2 and compressed StyleGAN2 on the LSUN-CAT dataset. (a), (b), (c), (d), (e) and (f) represent the full model, GS, CAGC, ITGC, \texttt{DiME}, and \texttt{NICKEL\&DiME}, respectively.}
\end{figure}

\clearpage

\bibliographystyle{splncs04}
\bibliography{egbib}